\documentclass[runningheads]{llncs}
\usepackage{chapterbib}
\usepackage{eccv}

\usepackage{eccvabbrv}
\usepackage{pdfpages}
\usepackage{graphicx}
\usepackage{booktabs}
\usepackage{multirow}
\usepackage{pifont}
\usepackage{marvosym}
\usepackage{algorithm}
\usepackage{algpseudocode}
\usepackage[accsupp]{axessibility}  % Improves PDF readability for those with disabilities.
\usepackage{hyperref}
\usepackage{orcidlink}

\begin{document}
\title{A Unified Anomaly Synthesis Strategy with Gradient Ascent for Industrial Anomaly Detection and Localization}
\titlerunning{A Unified Anomaly Synthesis Strategy}

\author{Qiyu Chen\inst{1,2}\orcidlink{0009-0003-7910-862X} \and
Huiyuan Luo\inst{1}\orcidlink{0000-0001-7855-9902} \and
Chengkan Lv\inst{1}\textsuperscript{(\Letter)}\orcidlink{0000-0001-5319-1363} \and
Zhengtao Zhang\inst{1,2,3}\orcidlink{0000-0003-1659-7879}}
\authorrunning{Q.~Chen et al.}

\institute{
Institute of Automation, Chinese Academy of Sciences, Beijing, China \and
School of Artificial Intelligence, University of Chinese Academy of Sciences \and
CASI Vision Technology CO., LTD., Luoyang, China \\
\email{\{chenqiyu2021,huiyuan.luo,chengkan.lv,zhengtao.zhang\}@ia.ac.cn}}

\maketitle

\begin{abstract}
Anomaly synthesis strategies can effectively enhance unsupervised anomaly detection. 
However, existing strategies have limitations in the coverage and controllability of anomaly synthesis,
particularly for weak defects that are very similar to normal regions.
In this paper, we propose Global and Local Anomaly co-Synthesis Strategy (GLASS),
a novel unified framework designed to synthesize a broader coverage of anomalies under the manifold and hypersphere distribution constraints of
Global Anomaly Synthesis (GAS) at the feature level and Local Anomaly Synthesis (LAS) at the image level.
Our method synthesizes near-in-distribution anomalies in a controllable way using Gaussian noise guided by gradient ascent and truncated projection.
GLASS achieves state-of-the-art results on the MVTec AD (detection AUROC of 99.9\%), VisA, and MPDD datasets and excels in weak defect detection.
The effectiveness and efficiency have been further validated in industrial applications for woven fabric defect detection.
The code and dataset are available at: \url{https://github.com/cqylunlun/GLASS}.
\end{abstract}
\section{Introduction}
\label{sec:intro}

\begin{figure*}[t]
  \centering
  \includegraphics[width=1\linewidth]{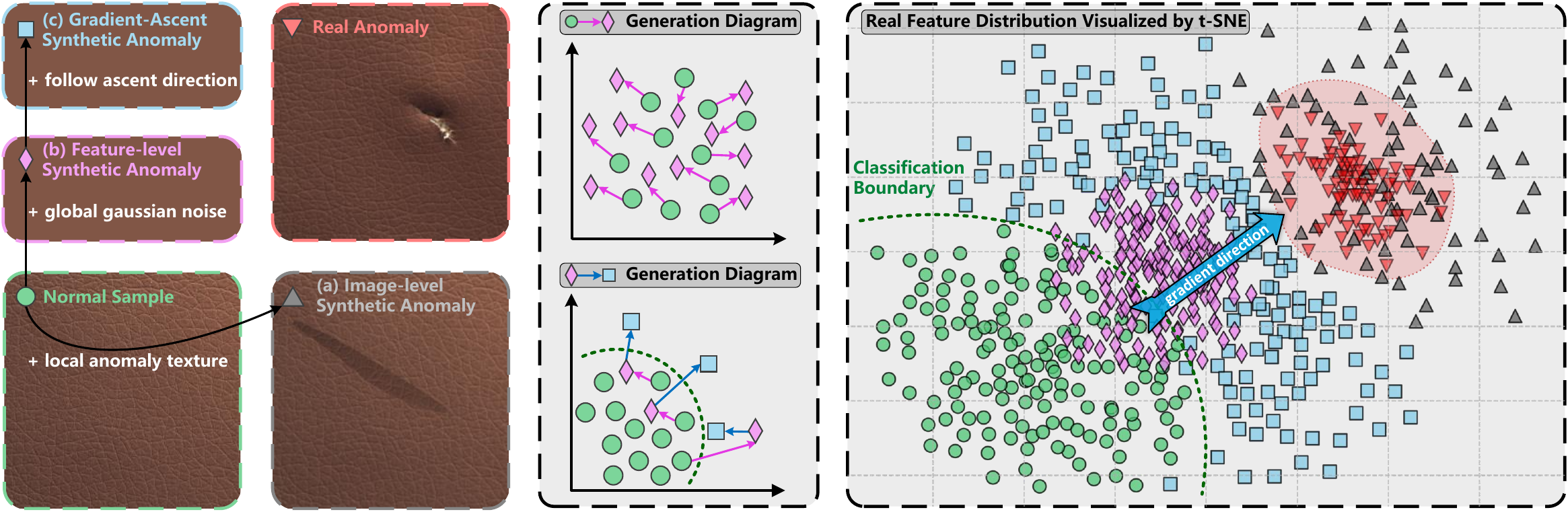}
  \caption{Process flow and visualization of various anomaly synthesis strategies.
  (a) Image-level anomaly synthesis strategy (gray triangles) provides detailed textures but lacks diversity.
  (b) Feature-level anomaly synthesis strategy (pink diamonds) is more efficient but lacks directionality.
  (c) Our method (blue squares) controls the distribution of synthetic anomalies at image and feature levels by using gradient ascent.
  }
  \label{fig:t-SNE}
\end{figure*}

Anomaly detection and localization aim to identify and localize abnormal regions by leveraging normal samples. 
Due to the challenge of collecting sufficient defect samples and the high cost of pixel-level annotations,
supervised approaches become impractical in these contexts.
Consequently, unsupervised anomaly detection techniques are widely applied in
industrial inspection scenarios \cite{roth2022towards,zavrtanik2022dsr,bae2023pni,hyun2024reconpatch}.
Moreover, since the weak defects are anomalies with small areas or low contrast,
some abnormal regions may be in close proximity to normal regions.

Existing anomaly detection methods can broadly be classified into three main categories.
Reconstruction-based methods \cite{zavrtanik2021reconstruction,akcay2019ganomaly} detect anomalies by
analyzing the residual image before and after reconstruction.
Embedding-based methods \cite{deng2022anomaly,lee2022cfa} leverage pre-trained networks to extract and
compress features into a compact space, distinctly separating anomaly features from normal clusters within the feature space.
These two categories are directly trained on original normal samples.
However, they fail to resolve the aforementioned issue.
Synthesis-based methods \cite{zavrtanik2021draem,tien2023revisiting,liu2023simplenet,zavrtanik2022dsr} typically synthesize anomalies from normal samples,
introducing anomaly discrimination information into the detection model for enhanced performance.

A common paradigm is the image-level anomaly synthesis strategy \cite{zavrtanik2021draem,tien2023revisiting,cao2023collaborative},
as depicted in Fig.~\ref{fig:t-SNE}(a), which explicitly simulates anomalies at the image level.
Although the image-level anomaly synthesis provides detailed anomaly textures, it is considered as lacking diversity and realism.
Recent methods \cite{zavrtanik2022dsr,liu2023simplenet,you2022unified} are based on the feature-level anomaly synthesis strategy,
as illustrated in Fig.~\ref{fig:t-SNE}(b), which implicitly simulates anomalies at the feature level.
Due to the smaller size of the feature maps, the feature-level anomaly synthesis is more efficient.
However, it also lacks the capability to synthesize anomalies directionally in a controllable way, particularly for near-in-distribution anomalies.

To address the limitations mentioned above, we propose Global and Local Anomaly co-Synthesis Strategy (GLASS),
a novel unified framework designed to synthesize a broader coverage of anomalies under the manifold and hypersphere distribution constraints of
Global Anomaly Synthesis (GAS) at the feature level and Local Anomaly Synthesis (LAS) at the image level.
Specifically, we propose the novel feature-level GAS, as illustrated in Fig.~\ref{fig:t-SNE}(c),
which utilizes Gaussian noise guided by gradient ascent and truncated projection. 
GAS synthesizes anomalies near the normal sample distribution in a controllable way,
resulting in a tighter classification boundary that further enhances weak defect detection.
The image-level LAS makes improvements by providing a more diverse range of anomaly synthesis. 
GAS synthesizes weak anomalies around normal points, while LAS synthesizes strong anomalies far from normal points.
Theoretically, the near-in-distribution anomalies synthesized by GAS are derived from normal features through relatively small noise and gradient ascents,
while the far-from-distribution anomalies synthesized by LAS are generated by significantly overlaying textures on normal images.
Therefore, the rightmost t-SNE visualization of Fig.~\ref{fig:t-SNE} shows that the anomalies guided by
gradient ascent predominantly position themselves close to the appropriate classification boundary.
Compared to the anomaly synthesis strategy based on Gaussian noise, our method minimizes the overlap between anomalous and normal samples,
reducing the risk of misclassifying normal samples as anomalies.

The main contributions of the proposed GLASS are summarised as follows:
\begin{itemize}
    \item We propose a unified framework for synthesizing a broader coverage of anomalies in a controllable way at image and feature levels.
    \item We propose a novel feature-level GAS method that utilizes Gaussian noise guided by gradient ascent to enhance weak defect detection.
    \item Extensive experiments demonstrate that GLASS outperforms state-of-the-art (SOTA) methods in industrial anomaly detection and localization tasks.
\end{itemize}
\section{Related Work}
\label{sec:related}

\noindent
\textbf{Reconstruction-based methods} such as AutoEncoders \cite{zavrtanik2021reconstruction,zhou2022rethinking},
detect anomalies by analyzing the residual image before and after reconstruction.
These methods assume that the model will properly reconstruct normal regions while failing to reconstruct abnormal regions.
However, they heavily rely on the quality of reconstructed image and face challenges with the difference analysis method.

\noindent
\textbf{Embedding-based methods} utilize pre-trained networks to extract features, subsequently compressing normal features into a compact space.
As a result, anomaly features are distinctly separated from normal clusters within the feature space.
Memory bank methods \cite{roth2022towards,bae2023pni,hou2021divide} archive representative normal features and detect anomalies through metric learning.
Similarly, one-class classification methods \cite{reiss2021panda,xiao2023restricted,lee2022cfa} further define explicit classification boundaries,
such as hyperplanes \cite{scholkopf2001estimating} or hyperspheres \cite{tax2004support}.
Normalizing flow \cite{dinh2017density} methods \cite{gudovskiy2022cflow,lei2023pyramidflow,yu2021fastflow} aim to transform the distribution of
normal samples into a standard Gaussian distribution, causing anomalies to exhibit low likelihood.
Knowledge distillation methods \cite{deng2022anomaly,salehi2021multiresolution,batzner2024efficientad} leverage the distinction in
anomaly detection capabilities between teacher and student networks.
Despite achieving good performance, these feature embedding methods are only trained on original normal samples, lacking the representation of anomaly samples.

\noindent
\textbf{Synthesis-based methods} view the synthesis of anomalies as a form of data augmentation from the normal samples. 
The objective is to introduce anomaly discrimination information and mitigate potential overfitting that may arise from mapping all normal samples to one point.
Most existing methods synthesize anomalies at the image level:
CutPaste \cite{li2021cutpaste} employs a straightforward approach by cutting normal regions and pasting them at random positions;
NSA \cite{schluter2022natural} uses Poisson image editing to seamlessly blend blocks of different sizes from various images,
synthesizing a series of anomalies that are more similar to natural sub-image irregularities; 
DRAEM \cite{zavrtanik2021draem} synthesizes anomalies by creating binary masks using Perlin noise and filling them with external textures in the normal images.
Recently, several methods synthesize anomalies in the feature space: 
DSR \cite{zavrtanik2022dsr} samples in quantified feature space and synthesizes weak defects through the similarity comparison of codebook feature vector; 
SimpleNet \cite{liu2023simplenet} and UniAD \cite{you2022unified} synthesize anomalies by adding Gaussian noise to the normal features.
Generally, image-level anomaly synthesis provides detailed anomaly textures but lacks diversity,
whereas feature-level anomaly synthesis is more efficient but faces challenges with directionality and controllability.

\begin{figure*}[t]
  \centering
  \includegraphics[width=1\linewidth]{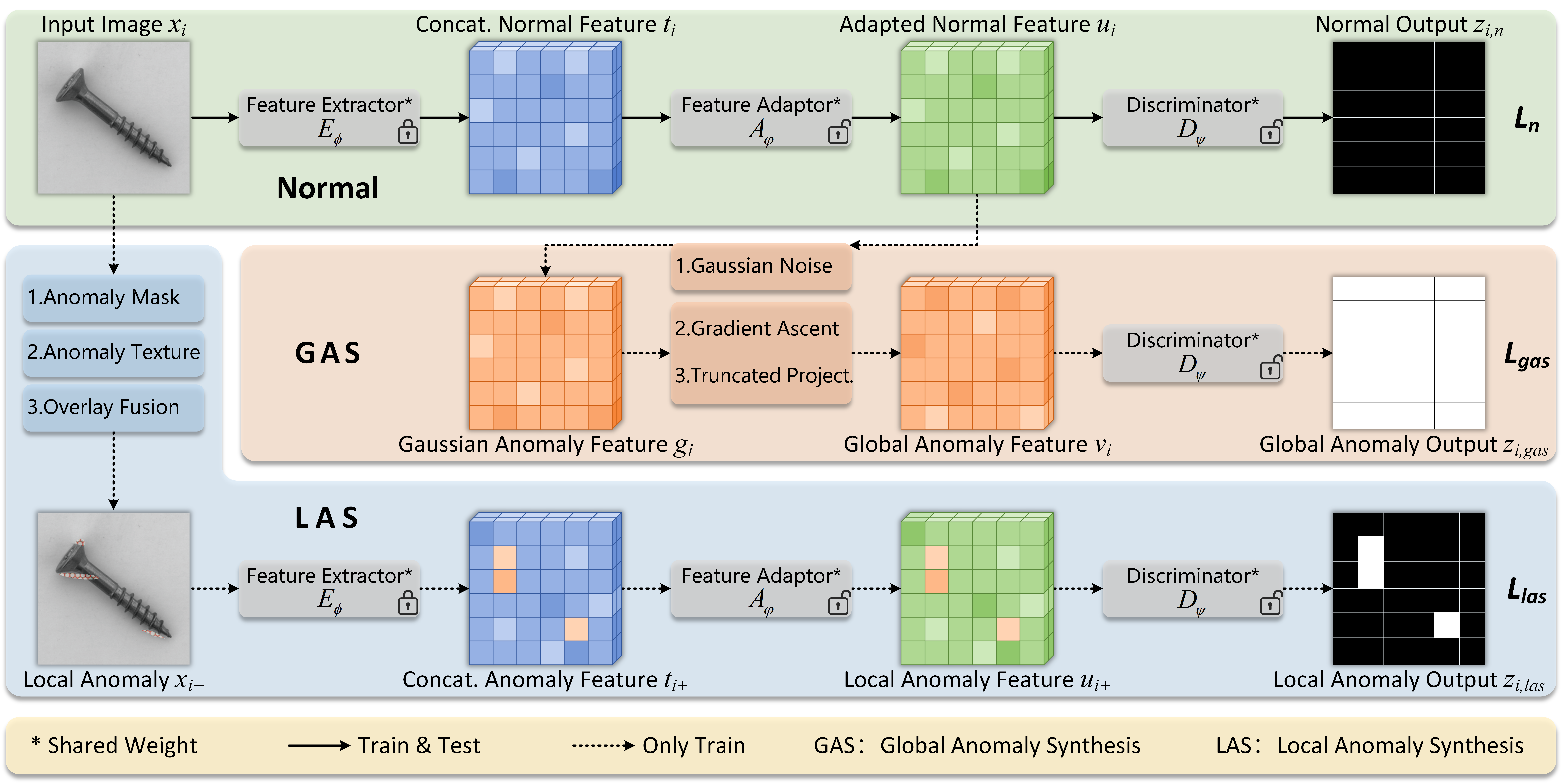}
  \caption{Schematic of the proposed GLASS. The training phase comprises three branches:
  (a) Normal branch obtains adapted normal features through a feature extractor and a feature adaptor.
  (b) GAS branch synthesizes global anomaly features from normal features in three steps based on gradient guidance.
  (c) LAS branch synthesizes local anomaly images from normal images in three steps based on texture overlay.}
  \label{fig:overview}
\end{figure*}

% ------------------------------------------------------------------------------------
\section{Proposed Method}
\label{sec:method}
The overall architecture of the proposed GLASS is shown in Fig.~\ref{fig:overview}.
During the training stage, GLASS primarily consists of three branches: Normal branch, GAS branch, and LAS branch.
Each branch shares three modules: a feature extractor \({{E}_{\phi }}\), a feature adaptor \({{A}_{\varphi }}\), and a discriminator \({{D}_{\psi }}\).
Normal samples are first processed by the frozen \({{E}_{\phi }}\) and the trainable \({{A}_{\varphi }}\) to obtain adapted normal features in Normal branch.
Next, global anomaly features are synthesized from the adapted normal features using gradient guidance in GAS branch.
Meanwhile, local anomaly images are synthesized by LAS branch through texture overlay,
which are then processed by \({{E}_{\phi }}\) and \({{A}_{\varphi }}\) to obtain local anomaly features.
Finally, the three features from the three branches are jointly fed into the discriminator \({{D}_{\psi }}\),
which is a segmentation network trained end-to-end using three loss functions.
During the inference phase, only the framework of normal branch is used to process the test images.

% ------------------------------------------------------------------------------------
\subsection{Feature Extractor and Feature Adaptor}
\label{sec:_EandA}
Similar to \cite{liu2023simplenet,lee2022cfa}, we utilize \({{A}_{\varphi }}\) to mitigate latent domain bias brought by the frozen \({{E}_{\phi }}\).
The feature map for image \mbox{\(x_i \in X_\text{train}\)} of level \(j\) extracted by the pre-trained backbone \(\phi\) is denoted as
\mbox{\(\phi_{i,j} = \phi_j(x_i) \in \mathbb{R}^{H_j \times W_j \times C_j}\)}.
The vector at location \((h, w)\) is represented as \mbox{\(\phi_{i,j}^{h,w} \in \mathbb{R}^{C_j}\)}.
By aggregating the neighborhood features through adaptive average pooling, the locally aware vector \mbox{\(s_{i,j}^{h,w} \in \mathbb{R}^{C_j}\)}
is derived from the neighborhood features of \(\phi_{i,j}^{h,w}\) with neighborhood size \(p\).
The set of vectors \(s_{i,j}^{h,w}\) constitutes the feature map \(s_{i,j}\).
By upsampling and merging \(s_{i,j}\) from different levels, the concatenated feature map \mbox{\(t_{i} \in \mathbb{R}^{H_m \times W_m \times C}\)} is denoted as
\mbox{\({{t}_{i}}={{E}_{\phi }}({{x}_{i}})\)}, where the channel size \(C = \sum_{j \in J} C_j\). The adapted normal vector \(u_{i}^{h,w}\) is denoted as
\mbox{\(u_{i}^{h,w}={{A}_{\varphi }}(t_{i}^{h,w})\)}, where \({{A}_{\varphi }}\) employs a single-layer perceptron with the same number of nodes in the input and output layers.

\begin{figure*}[t]
  \centering
  \includegraphics[width=1\linewidth]{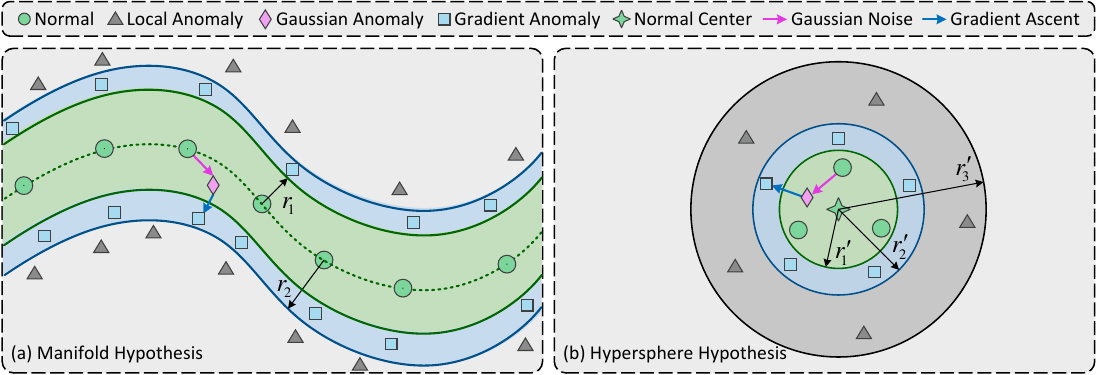}
  \caption{Schematic illustration of Global Anomaly Synthesis (GAS) under different hypotheses.
  Assume that \(r_m\)\ or \(r_h\) represents the \(L_{2}\) distance to manifold or hypersphere center, respectively.
  Green circles \mbox{(\(r_m < r_1\) or \(r_h < r'_1\))} represent normal features,
  gray triangles \mbox{(\(r_m > r_2\) or \(r'_2 < r_h < r'_3\))} represent local anomaly features,
  pink diamonds represent Gaussian anomaly features obtained by Gaussian noise from normal features,
  and blue squares \mbox{(\(r_1 < r_m < r_2\) or \(r'_1 < r_h < r'_2\))} represent global anomaly features
  obtained by gradient ascent and truncated projection from Gaussian anomaly features.}
  \label{fig:distribute}
\end{figure*}

% ------------------------------------------------------------------------------------
\subsection{Feature-level Global Anomaly Synthesis Strategy}
\label{sec:_gas}
Synthesizing anomalies in the feature space \cite{zavrtanik2022dsr,liu2023simplenet,you2022unified} has been proven to be an effective method.
However, existing methods lack the capability to synthesize anomalies directionally in a controllable way, particularly for near-in-distribution anomalies.
To more efficiently synthesize anomalies in feature space, we obtain global anomaly features by
adding Gaussian noise to normal features and constraining the synthetic direction of these anomalies using gradient ascent.
Here, ``global'' implies that anomalies are synthesized across all points of the feature map.
To avoid the excessive fluctuation of gradient ascent and make the anomaly synthesis more controllable,
truncated projection is employed to limit the minimum and maximum range of gradient ascent. The GAS is described as follows:

\noindent
\textbf{Distribution Hypothesis.}
It is posited that all normal feature points conform to either a manifold or a hypersphere distribution hypothesis \cite{pless2009survey}.
The manifold hypothesis assumes that the set of all normal feature points \(u_{i}^{h,w}\), denoted by \(U \subseteq \mathbb{R}^C\),
satisfies a low-dimensional locally linear manifold distribution \cite{goyal2020drocc}.
Since manifolds are locally linear and homeomorphic to Euclidean space,
the linear combination of low-dimensional embeddings can represent the global nonlinear distribution.
Under the manifold hypothesis illustrated in Fig.~\ref{fig:distribute}(a),
the feature set \mbox{\(N_a = \{ \tilde{u}_{i}^{h,w} \mid \|\tilde{u}_{i}^{h,w} - u_{j}^{h,w}\|_2 > r_1, \forall u_{j}^{h,w} \in U \}\)} is considered as anomalous.
The hypersphere hypothesis assumes that the set of \(u_{i}^{h,w}\) can be encompassed by a compact hypersphere \cite{ruff2018deep}.
Under the hypersphere hypothesis illustrated in Fig.~\ref{fig:distribute}(b),
the feature set \(N'_a = \{ \tilde{u}_{i}^{h,w} \mid \|\tilde{u}_{i}^{h,w} - c\|_2 > r'_1 \}\) is considered as anomalous,
where the center of hypersphere is defined as \(c = \frac{1}{|U|} \sum_{u_{i}^{h,w} \in U} u_{i}^{h,w}\).

Under the manifold and hypersphere hypothesis, the proposed GAS adopts a three-step method involving Gaussian noise, gradient ascent,
and truncated projection to synthesize global anomaly features. The first two steps of GAS are the same for manifold and hypersphere hypotheses.

\noindent
\textbf{Gaussian Noise.}
In real-world industrial settings, the distribution of anomalies is unknown.
Similar to \cite{liu2023simplenet,you2022unified}, Gaussian noise is adopted to simulate these diverse anomalies.
Specifically, the Gaussian anomaly feature point \( g_{i}^{h,w}\) is obtained by the addition of \(u_{i}^{h,w}\)
and noise \mbox{\(\varepsilon _{i}^{h,w} \sim N({{\mu }_{g}},\sigma _{g}^{2})\)}, denoted as \( g_{i}^{h,w} = u_{i}^{h,w} + \varepsilon_{i}^{h,w} \).
However, these Gaussian anomaly feature points are synthesized in an undirected way, leading to ineffective training for detection.

\noindent
\textbf{Gradient Ascent.} The most effective way to synthesize anomalies in feature space is to follow the direction of gradient ascent.
Leveraging the previously mentioned Gaussian noise, we integrate gradient information guided by the GAS branch loss \(L_{gas}\) in Eq.~\ref{eq:l_gas}.
We normalize the gradient vector and employ a learning rate \( \eta \) for the iterative acquisition of gradient anomaly feature \(\tilde{g}_{i}^{h,w}\) as:
\begin{equation}
  \tilde{g}_{i}^{h,w}=g_{i}^{h,w}+\eta \frac{\nabla {{L}_{gas}}(g_{i}^{h,w})}{\left\| \nabla {{L}_{gas}}(g_{i}^{h,w}) \right\|}
  \label{eq:gradient}
\end{equation}

\noindent
\textbf{Truncated Projection (Manifold).} Although \(\tilde{g}_{i}^{h,w}\) is derived from adding Gaussian noise to normal feature \(u_{i}^{h,w}\)
and guided by gradient ascent, there remains a risk of it being either too far from or too close to the normal feature.
Therefore, we propose truncated projection to constrain the range of gradient ascent, facilitating controllable anomaly synthesis.
The gradient ascent distance is calculated by \mbox{\(\tilde{\varepsilon}_{i}^{h,w} = \tilde{g}_{i}^{h,w} - u_{i}^{h,w}\)}.
To project \(\tilde{g}_{i}^{h,w}\) onto the set \(N_p = \{ \tilde{g}_{i}^{h,w} \mid r_1 < \|\tilde{g}_{i}^{h,w} - u_{i}^{h,w}\|_2 < r_2 \}\)
in Fig.~\ref{fig:distribute}(a), the truncated distance \(\hat{\varepsilon}_{i}^{h,w}\) is given by: 
\begin{equation}
  \hat{\varepsilon }_{i}^{h,w}=\frac{{{\alpha }_{i}}}{\left\| \tilde{\varepsilon }_{i}^{h,w} \right\|}
  \tilde{\varepsilon }_{i}^{h,w},\text{where }{{\alpha }_{i}}=\left\{ \begin{array}{l@{\hspace{4mm}}l}
   {{r}_{1}} & \left\| \tilde{\varepsilon }_{i}^{h,w} \right\|<{{r}_{1}}  \\[6pt]
   {{r}_{2}} & \left\| \tilde{\varepsilon }_{i}^{h,w} \right\|>{{r}_{2}}  \\[6pt]
   \left\| \tilde{\varepsilon }_{i}^{h,w} \right\| & \text{otherwise}  \\
  \end{array} \right.
  \label{eq:project_index}
\end{equation}
where the truncated coefficient \(\alpha_{i}\) depends on the magnitude of gradient ascent distance \(\|\tilde{\varepsilon}_{i}^{h,w}\|\). The manifold distance \( r_{1} \) and \( r_{2} \) are constants, typically \( r_2 = 2r_1 \). Finally, the global anomaly feature \( v_{i}^{h,w} = u_{i}^{h,w} + \hat{\varepsilon}_{i}^{h,w} \) is obtained. GAS algorithm under manifold hypothesis is presented in Alg.~\ref{alg:gas}.

\begin{algorithm}[t]
\caption{GAS under Manifold Hypothesis}
\begin{algorithmic}[1]
\State \textbf{Input:} normal feature map \(u_i\), number of batch \(n_{\text{batch}}\), number of iteration \(n_{\text{step}}\), interval of projection \(n_{\text{proj}}\)
\State \textbf{Output:} global anomaly feature map \( v_i \)
\For{\( \text{batch} = 1 \) to \( n_{\text{batch}} \)}
    \State Initialize \( u_i \) by \({{E}_{\phi }}\) and \({{A}_{\varphi }}\)
    \State \textit{Gaussian noise}. Add \( \varepsilon_i \) to \( u_i \rightarrow g_i \)
    \For{\( \text{step} = 1 \) to \( n_{\text{step}} \)}
        \State \textit{Gradient ascent}.
        \State (a) Calculate the loss \( L_{\text{gas}} \) of GAS branch by \( g_i \)
        \State (b) Update \( \tilde{g}_i \) according to Eq.~\ref{eq:gradient} with no grad.
        \If{\( \text{step} \) is a multiple of \(n_{\text{proj}}\)}
            \State \textit{Truncated projection}.
            \State (c) Get gradient ascent distance \( \tilde{\varepsilon}_i = \tilde{g}_i - u_i \)
            \State (d) Constrain the range by Eq.~\ref{eq:project_index} to get truncated distance \( \tilde{\varepsilon}_i \rightarrow \hat{\varepsilon}_i \)
            \State (e) Get GAS feature \( v_i = u_i + \hat{\varepsilon}_i \)
        \EndIf
    \EndFor
\EndFor
\State \textbf{return} \( v_i \)
\end{algorithmic}
\label{alg:gas}
\end{algorithm}

\begin{figure*}[t]
  \centering
  \includegraphics[width=1\linewidth]{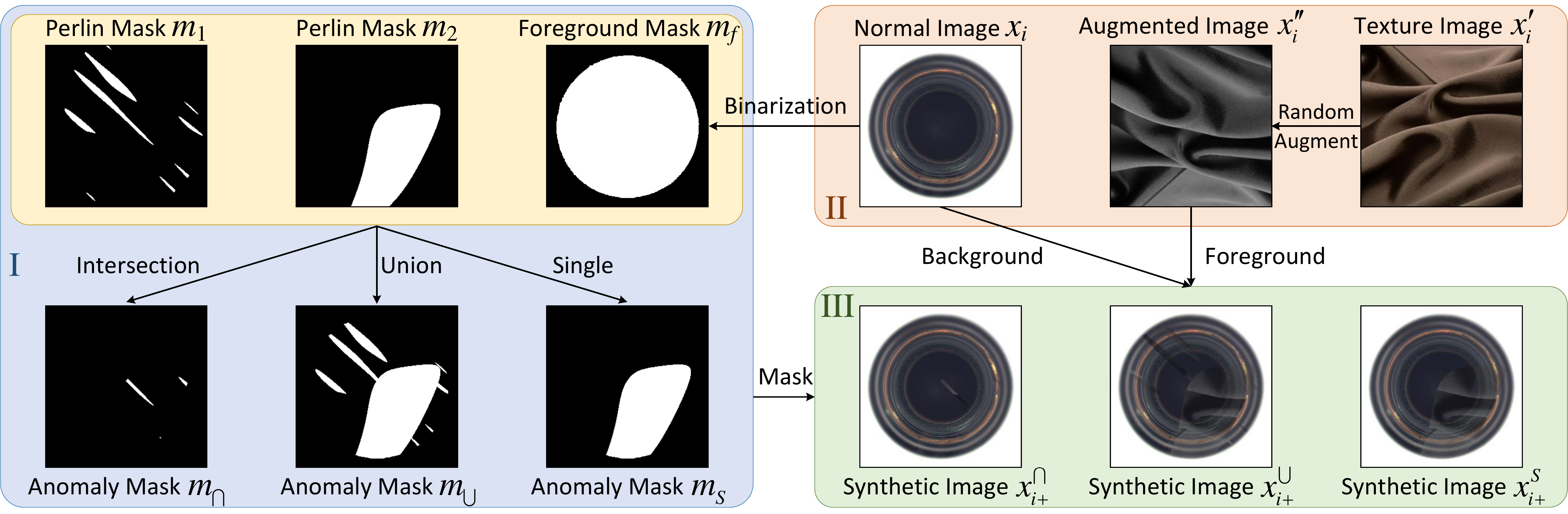}
  \caption{Flowchart of Local Anomaly Synthesis (LAS) consisting of three steps: Step I: Anomaly Mask, Step II: Anomaly Texture, and Step III: Overlay Fusion.}
  \label{fig:las}
\end{figure*}

\noindent
\textbf{Truncated Projection (Hypersphere).} Hypersphere hypothesis further constraints the distribution of
gradient anomaly features \(\tilde{g}_{i}^{h,w}\) from GAS and local anomaly features \( u_{i+}^{h,w} \) from LAS.
Similar to Eq.~\ref{eq:project_index}, global anomaly feature \( \tilde{v}_{i}^{h,w} \) is obtained by
projecting \(\tilde{g}_{i}^{h,w}\) onto the set \mbox{\( N'_p = \{ \tilde{g}_{i}^{h,w} \mid r'_1 < \|\tilde{g}_{i}^{h,w} - c\|_2 < r'_2 \} \)}.
Since \( u_{i+}^{h,w} \) is generally further away from the normal feature \( u_{i}^{h,w} \) than \( \tilde{v}_{i}^{h,w} \),
it is also projected onto the set \mbox{\( N''_p = \{ u_{i+}^{h,w} \mid r'_2 < \|u_{i+}^{h,w} - c\|_2 < r'_3 \} \)} in Fig.~\ref{fig:distribute}(b).
This is because \( u_{i+}^{h,w} \) is unlikely to merge with \( u_{i}^{h,w} \) after truncated projection,
which is a problem that might occur under the manifold hypothesis.
To make the normal samples more compact, the lower bound threshold \(r'_1\) denotes the radius of hypersphere,
which is iteratively updated and empirically set to cover 75\% of the normal samples.
This prevents synthetic anomalies from being too close to the center.
The upper bound threshold is typically set as \( r'_3 = 2r'_2 = 4r'_1 \).

Given the complex nonlinear structure of manifold distribution,
we posit that a more concentrated intraclass distribution aligns more closely with hypersphere distribution, and vice versa.
It is confirmed by the experiments that manifold distribution performs slightly better than
hypersphere distribution due to the complex nonlinear structures of most defects. 
In practice, we analyze the image-level spectrogram to determine the distribution hypothesis of different categories.
Details for the choice of hypothesis are provided in Sec.~{B} of the appendix.

% ------------------------------------------------------------------------------------
\subsection{Image-level Local Anomaly Synthesis Strategy}
\label{sec:_las}

Synthesizing anomalies within local regions can provide detailed anomaly textures.
Fusing DTD textures with Perlin masks to synthesize anomalies at the image level
is a commonly used approach in anomaly detection \cite{zavrtanik2021draem,zhang2023destseg,yang2023memseg}.
Building on this approach, we propose the image-level LAS to synthesize a more diverse range of anomalies.
Fig.~\ref{fig:las} presents the flowchart of LAS, detailed as follows: 

\noindent
\textbf{Anomaly Mask.} We first generate two binary masks by Perlin noise, denoted as \( m_1 \) and \( m_2 \).
Since anomalies generally appear on the surface of industrial samples,
the foreground mask \( m_f \) of normal sample is obtained through binarization inspired by \cite{yao2023explicit}.
To increase the diversity of anomalous regions, the intersection and union of \( m_1 \) and \( m_2 \) is utilized to construct the final mask \( {m}_{i} \) as:
\begin{equation}
  {{m}_{i}}=\left\{ \begin{array}{l@{\hspace{8mm}}l}
   ({{m}_{1}}\wedge {{m}_{2}})\wedge {{m}_{f}} & 0\le {{p}_{m}}\le \alpha   \\
   ({{m}_{1}}\vee {{m}_{2}})\wedge {{m}_{f}} & \alpha <{{p}_{m}}\le 2\alpha   \\
   {{m}_{1}}\wedge {{m}_{f}} & 2\alpha <{{p}_{m}}\le 1  \\
  \end{array} \right.
  \label{eq:mask}
\end{equation}
where random number \(p_m \sim U(0,1)\), with \(\alpha\) set to \(\frac{1}{3}\) in the experiments.

\noindent
\textbf{Anomaly Texture.} After determining the shape of the anomalous region,
we randomly select an image \(x'_i\) from the texture dataset DTD \cite{cimpoi2014describing}.
From the set of \(K=9\) image augmentation methods \(T = \{T_1, \ldots, T_K\}\),
we randomly choose three methods to form \(T_R \subset T\), similar to RandAugment \cite{cubuk2020randaugment}.
The augmented anomaly texture image is then obtained as \(x''_i = T_R(x'_i)\).

\noindent
\textbf{Overlay Fusion.} To better simulate weak defects while covering more detailed anomalies,
we adopt the transparency coefficient \(\beta \sim N(\mu_m, \sigma^2_m)\) to modulate the proportion of training set image \(x_i\)
within the synthetic image under the anomaly mask. The local anomaly image \({x}_{i+}\) is fused as:
\begin{equation}
  {{x}_{i+}}={{x}_{i}}\odot {\bar{m}_{i}}+(1-\beta ){x''_i}\odot {{m}_{i}}+\beta {{x}_{i}}\odot {m}_{i}
  \label{eq:fusion}
\end{equation}
where \({\bar{m}_{i}}\) is derived by inverting the anomaly mask \({m}_{i}\).
Subsequently, \(x_{i+}\) is processed through Sec.~\ref{sec:_EandA} to obtain
the local anomaly feature map \({{u}_{i+}}={{A}_{\varphi }}({{E}_{\phi }}({{x}_{i+}}))\),
with the corresponding position denoted as \(u_{i+}^{h,w}\) at \((h,w)\).

% ------------------------------------------------------------------------------------
\subsection{Discriminator and Training Objectives}

Three groups of features are obtained through three branches, respectively, serving as the input for the discriminator \({{D}_{\psi }}\).
It employs a single hidden layer MLP with Sigmoid, directly outputting the anomaly confidence \(z_i^{h,w} \in \mathbb{R}\) for each feature point.
The training objectives typically consist of three components.

The first term \(L_n\) is given by the Binary Cross-Entropy (BCE) loss between the normal feature discrimination
\mbox{\(z_{i,n}={{D}_{\psi }}(u_{i})\)} and the ground truth of full-size feature map normal:
\begin{equation}
  {{L}_{n}}=\sum\limits_{{{x}_{i}}\in {{X}_{train}}}{{{f}_{BCE}}({{z}_{i,n}},\mathbf{0})}
  \label{eq:l_n}
\end{equation}

The second term \(L_{gas}\) is given by the BCE loss between the global anomaly feature discrimination
\mbox{\(z_{i,gas}={{D}_{\psi }}(v_{i})\)} and the ground truth of full-size feature map anomaly:
\begin{equation}
  {{L}_{gas}}=\sum\limits_{{{x}_{i}}\in {{X}_{train}}}{{{f}_{BCE}}({{z}_{i,gas}},\mathbf{1})}
  \label{eq:l_gas}
\end{equation}

To address the imbalance issue in binary classification of the local anomaly features,
the third term \(L_{las}\) is given by the Focal loss \cite{lin2017focal} between the local anomaly feature discrimination
\mbox{\(z_{i,las}={{D}_{\psi }}(u_{i+})\)} and the ground truth of anomaly mask \({m}_{i}\):
\begin{equation}
  {{L}_{las}}=\sum\limits_{{{x}_{i}}\in {{X}_{train}}}{{{f}_{Focal}}({{z}_{i,las}},{{m}_{i}})}
  \label{eq:l_las}
\end{equation}

To filter crucial samples such as weak defects,
Online Hard Example Mining (OHEM) \cite{shrivastava2016training} is applied to \(L_{las}\). The overall loss function is:
\begin{equation}
  L={{L}_{n}}+{{L}_{gas}}+{{f}_{ohem}}({{L}_{las}})
  \label{eq:l_l_l}
\end{equation}

% ------------------------------------------------------------------------------------
\subsection{Inference and Anomaly Scoring}

As depicted in Fig.~\ref{fig:overview}, the inference process is represented by the solid line without GAS and LAS.
Input image \mbox{\(x_i \in X_\text{test}\)} is processed by Sec.~\ref{sec:_EandA} to obtain \mbox{\({{u}_{i}}={{A}_{\varphi }}({{E}_{\phi }}({{x}_{i}}))\)}.
Subsequently, \({{D}_{\psi }}\) gives the segmentation result \mbox{\(z_i = D_\psi(u_i)\)}.
By upsampling the interpolation of \mbox{\(z_i \in \mathbb{R}^{H_m \times W_m}\)} to the original image size and applying Gaussian smoothing to mitigate noise,
the pixel-level anomaly score \(S_{AL}\) used for anomaly localization is obtained as:
\begin{equation}
  {{S}_{AL}}={{f}_{smooth}}(f_{resize}^{{{H}_{0}},{{W}_{0}}}({{z}_{i}}))
  \label{eq:score}
\end{equation}

Additionally, the image-level anomaly score \(S_{AD}\) used for anomaly detection is given by the maximum value of all points in \(z_i\).
\section{Experiments}
\label{sec:exper}

% Table generated by Excel2LaTeX from sheet 'Table 1'
\begin{table*}[t]\small
  \centering
  \caption{Comparison of GLASS and its variants with different SOTA methods on each category of MVTec AD. ·/· denotes image-level AUROC\% and pixel-level AUROC\%.}
  \resizebox{0.96\textwidth}{!}{
    \begin{tabular}{l@{\hspace{10pt}}|c@{\hspace{10pt}}c@{\hspace{10pt}}c@{\hspace{10pt}}c@{\hspace{10pt}}c@{\hspace{10pt}}|c@{\hspace{10pt}}c@{\hspace{10pt}}c}
    \hline
    Category & DSR   & PatchCore & BGAD & RD++  & SimpleNet & GLASS-m & GLASS-h & GLASS-j \\
    \hline
    Carpet & 99.6/96.0 & 98.6/99.1 & 99.8/99.4 & \textbf{100}/99.2 & 99.7/98.4 & 99.8/\textbf{99.6} & 99.2/99.4 & 99.8/\textbf{99.6} \\
    Grid  & \textbf{100/99.6} & 97.7/98.8 & 99.1/99.4 & \textbf{100}/99.3 & 99.9/98.5 & \textbf{100}/99.4 & \textbf{100}/99.0 & \textbf{100}/99.4 \\
    Leather & 99.3/99.5 & \textbf{100}/99.3 & \textbf{100}/99.7 & \textbf{100}/99.5 & \textbf{100}/99.2 & \textbf{100/99.8} & \textbf{100/99.8} & \textbf{100/99.8} \\
    Tile  & \textbf{100}/98.6 & 98.8/95.7 & \textbf{100}/96.7 & 99.7/96.6 & 98.7/97.7 & \textbf{100/99.7} & \textbf{100}/99.1 & \textbf{100/99.7} \\
    Wood  & 94.7/91.5 & 99.1/95.0 & 99.5/97.0 & 99.3/95.8 & 99.5/94.4 & \textbf{99.9/98.8} & 99.7/97.6 & \textbf{99.9/98.8} \\
    \hline
    Texture Avg. & 98.7/97.0 & 98.9/97.6 & 99.7/98.4 & 99.8/98.1 & 99.6/97.6 & \textbf{99.9/99.5} & 99.8/99.0 & \textbf{99.9/99.5} \\
    \hline
    Bottle & 99.6/98.8 & \textbf{100}/98.5 & \textbf{100}/98.9 & \textbf{100}/98.8 & \textbf{100}/98.0 & \textbf{100}/99.2 & \textbf{100/99.3} & \textbf{100/99.3} \\
    Cable & 95.3/97.7 & 99.8/98.4 & 97.9/98.0 & 99.3/98.4 & \textbf{100}/97.5 & 98.2/98.1 & 99.8/\textbf{98.7} & 99.8/\textbf{98.7} \\
    Capsule & 98.3/91.0 & 98.1/99.0 & 97.3/99.1 & 99.0/98.8 & 97.8/98.9 & \textbf{99.9/99.4} & 99.8/99.3 & \textbf{99.9/99.4} \\
    Hazelnut & 97.7/99.1 & \textbf{100}/98.7 & 99.3/98.5 & \textbf{100}/99.2 & 99.8/98.1 & \textbf{100/99.4} & \textbf{100}/99.1 & \textbf{100/99.4} \\
    Metal nut & 99.1/94.1 & \textbf{100}/98.3 & 99.3/97.7 & \textbf{100}/98.1 & \textbf{100}/98.8 & \textbf{100/99.4} & \textbf{100}/99.1 & \textbf{100/99.4} \\
    Pill  & 98.9/94.2 & 96.4/97.8 & 98.8/98.0 & 98.4/98.3 & 98.6/98.6 & \textbf{99.4/99.5} & 99.3/99.4 & 99.3/99.4 \\
    Screw & 95.9/98.1 & 98.4/99.5 & 92.3/99.2 & 98.9/99.7 & 98.7/99.2 & 99.5/\textbf{99.5} & \textbf{100/99.5} & \textbf{100/99.5} \\
    Toothbrush & \textbf{100/99.5} & \textbf{100}/98.6 & 86.9/98.7 & \textbf{100}/99.1 & \textbf{100}/98.5 & \textbf{100}/99.3 & \textbf{100}/99.0 & \textbf{100}/99.3 \\
    Transistor & 96.3/80.3 & 99.9/96.1 & 99.7/93.9 & 98.5/94.3 & \textbf{100}/97.0 & 99.0/95.5 & 99.9/\textbf{97.6} & 99.9/\textbf{97.6} \\
    Zipper & 98.5/98.4 & 99.4/98.9 & 97.8/98.7 & 98.6/98.8 & 99.9/98.9 & \textbf{100/99.6} & 99.9/99.3 & \textbf{100/99.6} \\
    \hline
    Object Avg. & 98.0/95.1 & 99.2/98.4 & 96.9/98.1 & 99.3/98.3 & 99.5/98.3 & 99.6/98.9 & \textbf{99.9}/99.0 & \textbf{99.9/99.2} \\
    \hline
    Average & 98.2/95.8 & 99.1/98.1 & 97.9/98.2 & 99.4/98.3 & 99.5/98.1 & 99.7/99.1 & 99.8/99.0 & \textbf{99.9/99.3} \\
    \hline
    \end{tabular}%
    }
  \label{tab:compare_mvtec}%
\end{table*}%

% ------------------------------------------------------------------------------------
\subsection{Datasets}
\label{sec:_dataset}

Three widely-used real-world public datasets are employed: MVTec AD \cite{bergmann2019mvtec}, VisA \cite{zou2022spot}, and MPDD \cite{jezek2021deep}.
Additionally, we construct the woven fabric defect detection (WFDD) dataset under industrial settings with 3860 normal and 241 anomaly samples.
To evaluate the ability of GLASS in weak defect detection, we create two test sets based on MVTec AD. MVTec AD-manual (MAD-man) consists of five subsets,
each constructed by one of five individuals who selected weak defect samples from every category of MVTec AD under unbiased conditions.
Due to the scarcity of weak defect, we also synthesize a weak defect test set named MVTec AD-synthesis (MAD-sys) from five texture categories of MVTec AD.
MAD-sys consists of four subsets with different levels of weakness, obtained by adjusting \mbox{\(\beta =\{0.1,0.3,0.5,0.7\}\)} in Eq.~\ref{eq:fusion}.
The WFDD, MAD-man, and MAD-sys datasets are released at this \href{https://github.com/cqylunlun/GLASS?tab=readme-ov-file#dataset-release}{website}.
Details of these datasets are provided in Sec.~A of the appendix.

% ------------------------------------------------------------------------------------
\subsection{Implementation Detail}

\noindent
\textbf{Experimental Settings.} We employ WideResnet50 as the backbone of \({{E}_{\phi }}\) and merge the features of level2 and level3 for GLASS.
The neighborhood size \(p\) is set to 3. Input images are resized and center cropped to \(288\times288\).
For LAS, transparency coefficient \(\beta \sim N(0.5, 0.1^2)\) is truncated within the range \([0.2,0.8]\).
For GAS, Gaussian noise \(\varepsilon \sim N(0,0.015^{2})\). GLASS-m is based on the manifold hypothesis where \(r_1=1, r_2=2\) in Eq.~\ref{eq:project_index}.
GLASS-h is based on the hypersphere hypothesis. GLASS-j is a hybrid strategy derived from judgment that integrates GLASS-h and GLASS-m.
The choice between GLASS-h and GLASS-m for each category is determined through the image-level spectrogram analysis method.
As the three variants of GLASS are highly similar, most experiments using GLASS-m by default.
We utilize the Adam optimizer to train \({{A}_{\varphi }}\) and \({{D}_{\psi }}\) with learning rates of 0.0001 and 0.0002, respectively.
The training epochs are set to 640 and the batch size is 8. All experiments are implemented
on an NVIDIA Tesla A800 GPU and an Intel(R) Xeon(R) Gold 6346 CPU @3.10GHz. 

\noindent
\textbf{Evaluation Metrics.} Area Under the Receiver Operating Characteristic Curve (AUROC) is a commonly used evaluation metric in anomaly detection,
we use it to evaluate the discriminative ability of models at image and pixel levels.
To provide a more comprehensive evaluation of the anomaly localization ability, we also calculate Per-Region Overlap (PRO) at pixel level. 

\begin{figure}[htbp]
  \centering
  \includegraphics[width=0.6\linewidth]{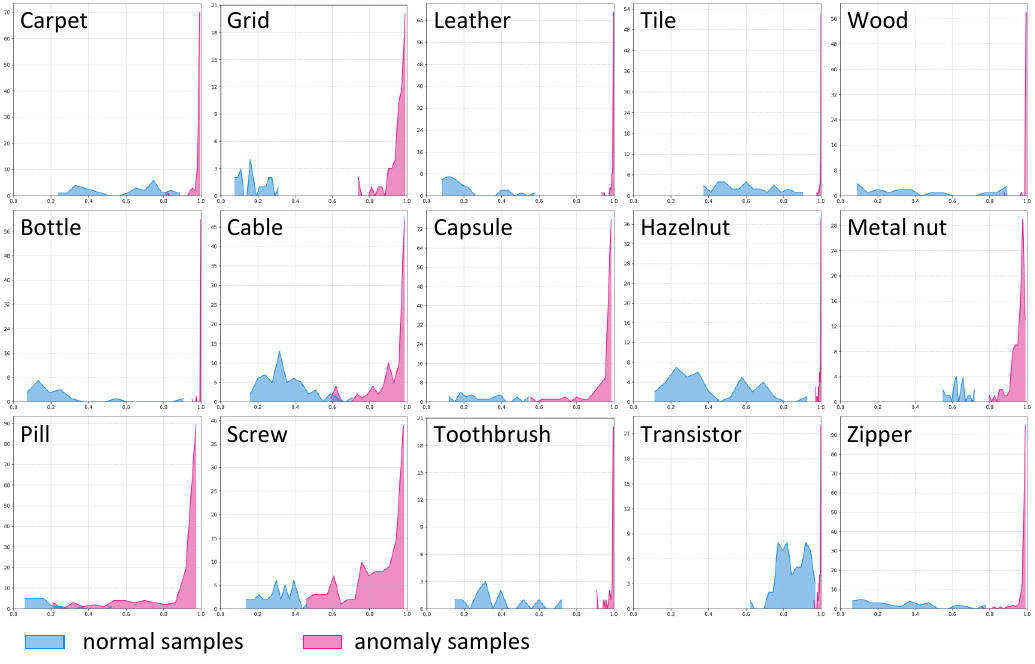}
  \caption{Anomaly score histograms of GLASS-j on each category of MVTec AD.}
  \label{fig:dataset_histogram}
\end{figure}

% ------------------------------------------------------------------------------------
\subsection{Comparative Experiments on Different Datasets}

According to \cite{liu2024deep}, five top SOTA methods across different subfields are employed,
including DSR \cite{zavrtanik2022dsr}, PatchCore \cite{roth2022towards}, BGAD\(\mathrm{}^{w/o}\) \cite{yao2023explicit}, RD++ \cite{tien2023revisiting},
and SimpleNet \cite{liu2023simplenet}. More comparative experiments are provided in Sec.~D of the appendix.

% Table generated by Excel2LaTeX from sheet 'Table 2'
\begin{table*}[htbp]\small
  \centering
  \caption{Comparison of GLASS and its variants with different SOTA methods on four datasets.
  ·/·/· denotes image-level AUROC\%, pixel-level AUROC\%, and pixel-level PRO\%. The last column provides the throughput measured in img/s.}
    \resizebox{1\textwidth}{!}{
    \begin{tabular}{l|c|c|c|c|c|c}
    \hline
    \multicolumn{1}{l|}{Method} & MVTec AD & VisA  & MPDD  & WFDD   & Avg.  & Throughput \\
    \hline
    DSR \cite{zavrtanik2022dsr} & 98.2/95.8/91.7 & 88.0/84.3/61.9 & 81.0/76.2/58.4 & 95.1/87.9/78.0 & 90.6/86.0/72.5 & 582 \\
    PatchCore \cite{roth2022towards} & 99.1/98.1/92.8 & 94.7/98.5/91.8 & 93.5/98.9/95.0 & 96.3/98.1/91.7 & 95.9/98.4/92.8 & 31 \\
    BGAD \cite{yao2023explicit} & 97.9/98.2/96.3 & 96.4/98.6/92.0 & 91.8/98.1/93.3 & 97.1/98.5/88.5 & 95.8/98.3/92.5 & 206 \\
    RD++ \cite{tien2023revisiting} & 99.4/98.3/95.0 & 96.3/98.7/92.2 & 95.5/98.7/95.6 & 95.2/98.4/92.9 & 96.6/98.5/93.9 & 623 \\
    SimpleNet \cite{liu2023simplenet} & 99.5/98.1/90.0 & 97.1/98.2/90.7 & 98.1/98.7/95.7 & 98.8/98.0/90.6 & 98.4/98.2/91.8 & 1306 \\
    \hline
    \multicolumn{1}{l|}{GLASS-m} & 99.7/99.1/96.4 & \textbf{98.8}/98.7/92.5 & \textbf{99.6/99.4/98.2} & \textbf{100/98.9/94.9} & 99.5/99.0/95.5 & \textbf{1327} \\
    \multicolumn{1}{l|}{GLASS-h} & 99.8/99.0/95.9 & 98.2/98.6/90.8 & 96.7/98.8/96.4 & 99.0/98.4/88.0 & 98.4/98.7/92.8 & \textbf{1327} \\
    \multicolumn{1}{l|}{GLASS-j} & \textbf{99.9/99.3/96.8} & \textbf{98.8/98.8/92.8} & \textbf{99.6/99.4/98.2} & \textbf{100/98.9/94.9} & \textbf{99.6/99.1/95.7} & \textbf{1327} \\
    \hline
    \end{tabular}%
    }
  \label{tab:compare_dataset}%
\end{table*}%

\begin{figure}[htbp]
  \centering
  \includegraphics[width=0.9\linewidth]{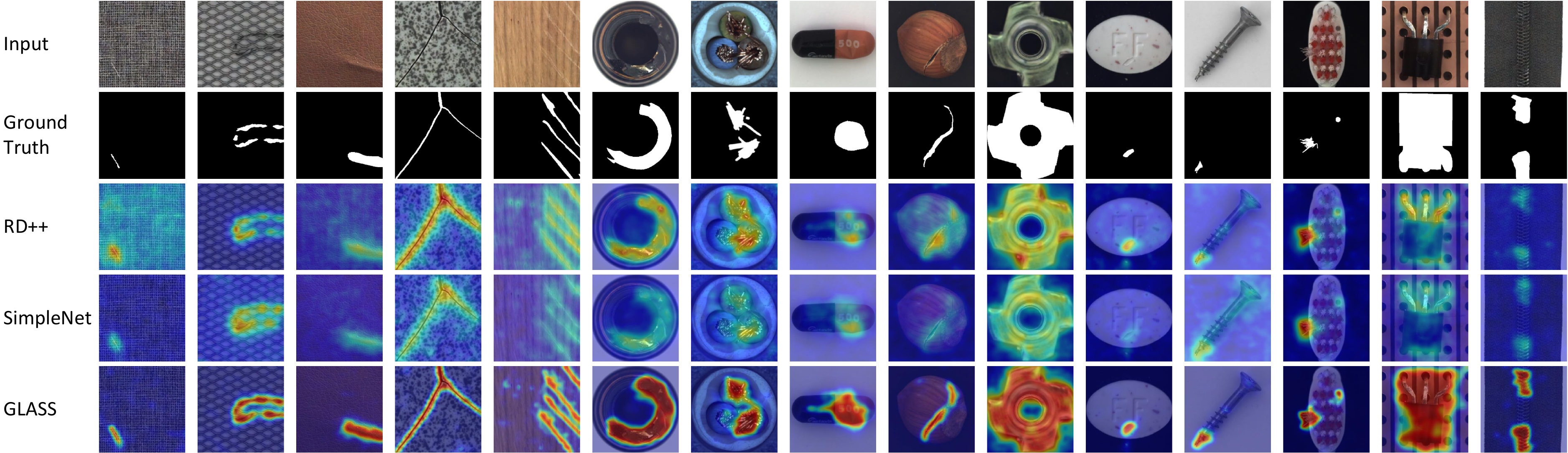}
  \caption{Qualitative results of GLASS-j and different SOTA methods on MVTec AD.}
  \label{fig:mvtec_visual}
\end{figure}

\begin{figure}[htbp]
  \centering
  \includegraphics[width=0.9\linewidth]{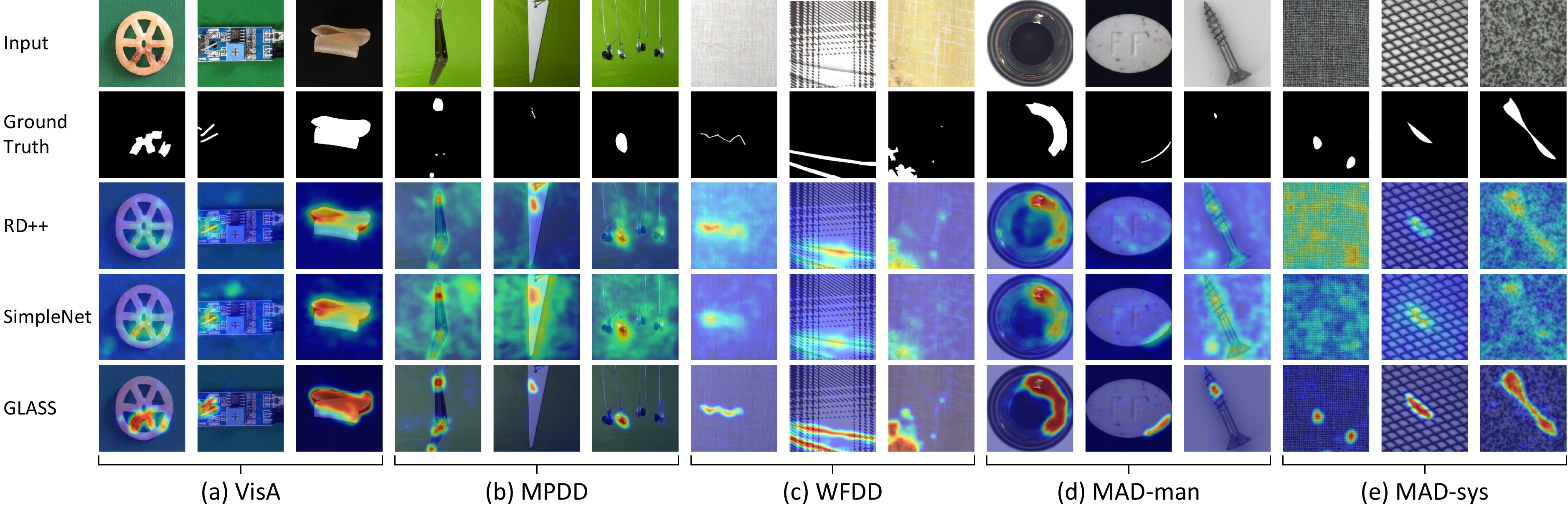}
  \caption{Qualitative results of GLASS-m and different SOTA methods on datasets.}
  \label{fig:weak_visual}
\end{figure}

\noindent
\textbf{Anomaly Detection on MVTec AD.} As shown in Tab.~\ref{tab:compare_mvtec}, GLASS-j achieves 100\% image-level AUROC on 9 out of 15 categories,
further achieving SOTA performance with an average of \textbf{99.9\%} image-level AUROC and \textbf{99.3\%} pixel-level AUROC on MVTec AD.
Specifically, GLASS-m excels in categories with complex nonlinear structures due to the locally-sensitive manifold distribution,
while GLASS-h based on hypersphere distribution is inclined towards categories with concentrated intraclass distribution.
The histograms in Fig.~\ref{fig:dataset_histogram} show a tiny overlap between normal and abnormal samples,
with the anomaly scores for anomalies being markedly high and concentrated.
Similarly, Fig.~\ref{fig:mvtec_visual} shows that GLASS-j has excellent discriminative ability between normal and abnormal samples. 

% Table generated by Excel2LaTeX from sheet 'Table 3'
\begin{table}[htbp]\small
  \centering
  \caption{Comparison of GLASS-m with different methods on two weak defect test sets.
  ·/·/· denotes image-level AUROC\%, pixel-level AUROC\%, and pixel-level PRO\%.}
  \resizebox{0.6\textwidth}{!}{
    \begin{tabular}{l|c|c|c}
    \hline
    \multicolumn{1}{l|}{Method} & MAD-man & MAD-sys & Avg. \\
   \hline
    DSR \cite{zavrtanik2022dsr}   & 94.2/96.9/91.3 & 91.5/90.2/78.3 & 92.9/93.5/84.8 \\
    PatchCore \cite{roth2022towards} & 97.6/98.6/94.7 & 92.4/92.6/63.9 & 95.0/95.6/79.3 \\
    BGAD \cite{yao2023explicit} & 95.9/98.7/96.1 & 90.4/86.8/64.0 & 93.2/92.8/80.0 \\
    RD++ \cite{tien2023revisiting}  & 98.2/98.9/96.7 & 83.9/86.3/61.8 & 91.1/92.6/79.2 \\
    SimpleNet \cite{liu2023simplenet} & 99.3/98.6/94.8 & 84.6/85.0/57.3 & 92.0/91.8/76.0 \\
    \hline
    GLASS & \textbf{99.6/99.3/97.5} & \textbf{95.6/93.3/80.3} & \textbf{97.6/96.3/88.9} \\
    \hline
    \end{tabular}%
    }
  \label{tab:compare_weak}%
\end{table}%

\noindent
\textbf{Anomaly Detection on Four Datasets.} Tab.~\ref{tab:compare_dataset} demonstrates that all three variants of GLASS outperform other SOTA methods
across the four datasets with higher speed. Compared to SimpleNet (based on feature-level anomaly synthesis),
GLASS-j increases the average image-level AUROC by 1.2\%, pixel-level AUROC by 0.9\%, and pixel-level PRO by 3.9\%.
With a simpler architecture, GLASS achieves superior precision and efficiency on the self-built dataset WFDD collected in industrial settings,
further confirming the feasibility of our method. 
As illustrated in Fig.~\ref{fig:weak_visual}(a-c), GLASS shows outstanding performance in detecting various types of anomalies across different industrial settings.

\noindent
\textbf{Anomaly Detection on Weak Defect.} Tab.~\ref{tab:compare_weak} shows the average performance of different methods on MAD-man and MAD-sys,
where GLASS surpasses all other methods significantly. Compared to SimpleNet, GLASS achieves improvements of 5.6\%, 4.5\%, and 12.9\% in three metrics,
surpassing the level of improvement observed on MVTec AD.
Fig.~\ref{fig:weak_visual}(d-e) presents the samples from MAD-man and MAD-sys, showing the outstanding performance of GLASS in weak defect detection.
More qualitative results are provided in Sec.~E of the appendix.

% ------------------------------------------------------------------------------------
\subsection{Ablation Study}
To verify the contribution of different modules, particularly in weak defect detection,
we have conducted corresponding ablation experiments mostly on MVTec AD. More ablation studies are provided in Sec.~C of the appendix.

\noindent
\textbf{Anomaly Synthesis Strategies.} We split GAS into three components: Gaussian Noise (GN), Gradient Ascent (GA), and Truncated Projection (TP).
As indicated in Tab.~\ref{tab:ablation_gas}, GAS (without GA and TP) performs better than LAS on MVTec AD.
This indicates that GAS has the advantage of detecting various types of anomalies.
However, LAS shows superior performance in weaker defects on MAD-sys, revealing its advantage in detecting local anomalies.
The cooperative training of LAS and GAS achieves an obvious improvement, showing their complementarity to synthesize a broader coverage of anomalies.

% Table generated by Excel2LaTeX from sheet 'Table 4'
\begin{table}[htbp]\small
  \centering
  \caption{Performance of GLASS-m on MVTec AD and two weak defect test sets under different anomaly synthesis strategies.
  ·/·/· denotes image-level AUROC\%, pixel-level AUROC\%, and pixel-level PRO\%.}
  \resizebox{0.63\textwidth}{!}{
    \begin{tabular}{c|c|c|c|c|c|c}
    \hline
    \multirow{2}{*}{LAS} & \multicolumn{3}{c|}{GAS} & \multirow{2}{*}{MVTec AD} & \multirow{2}{*}{MAD-man} & \multirow{2}{*}{MAD-sys} \\
    \cline{2-4}
        & GN    & GA    & TP    &       &       &  \\
    \hline
    \ding{51}     &       &       &       & 98.2/95.4/88.0 & 97.4/97.1/94.0 & 94.4/92.0/80.0 \\
          & \ding{51}     &       &       & 99.4/98.1/91.8 & 98.4/98.3/95.1 & 84.1/85.4/60.6 \\
    \ding{51}     & \ding{51}     &       &       & 99.5/98.9/94.7 & 98.7/99.1/96.8 & 94.6/92.2/77.7 \\
    \ding{51}     & \ding{51}     & \ding{51}     &       & 99.6/99.0/95.9 & 99.0/99.2/97.1 & 95.0/92.8/79.5 \\
    \ding{51}     & \ding{51}     & \ding{51}     & \ding{51}     & \textbf{99.7/99.1/96.4} & \textbf{99.6/99.3/97.5} & \textbf{95.6/93.3/80.3} \\
    \hline
    \end{tabular}
    }
  \label{tab:ablation_gas}
\end{table}

\noindent
\textbf{Components in GAS.} Tab.~\ref{tab:ablation_gas} explicitly shows that the three evaluation metrics improve successively by adding GN, GA, and TP. 
As LAS and GAS (without GA and TP) have already achieved 99.5\% image-level AUROC on MVTec AD, the introduction of GA and TP offers relative improvement. 
However, their improvements are more significant on MAD-man and MAD-sys, indicating that GA and TP are particularly effective in detecting weak defects.

\begin{figure}[t]
  \centering
  \begin{minipage}{0.55\textwidth}
    \centering
    \includegraphics[width=\linewidth]{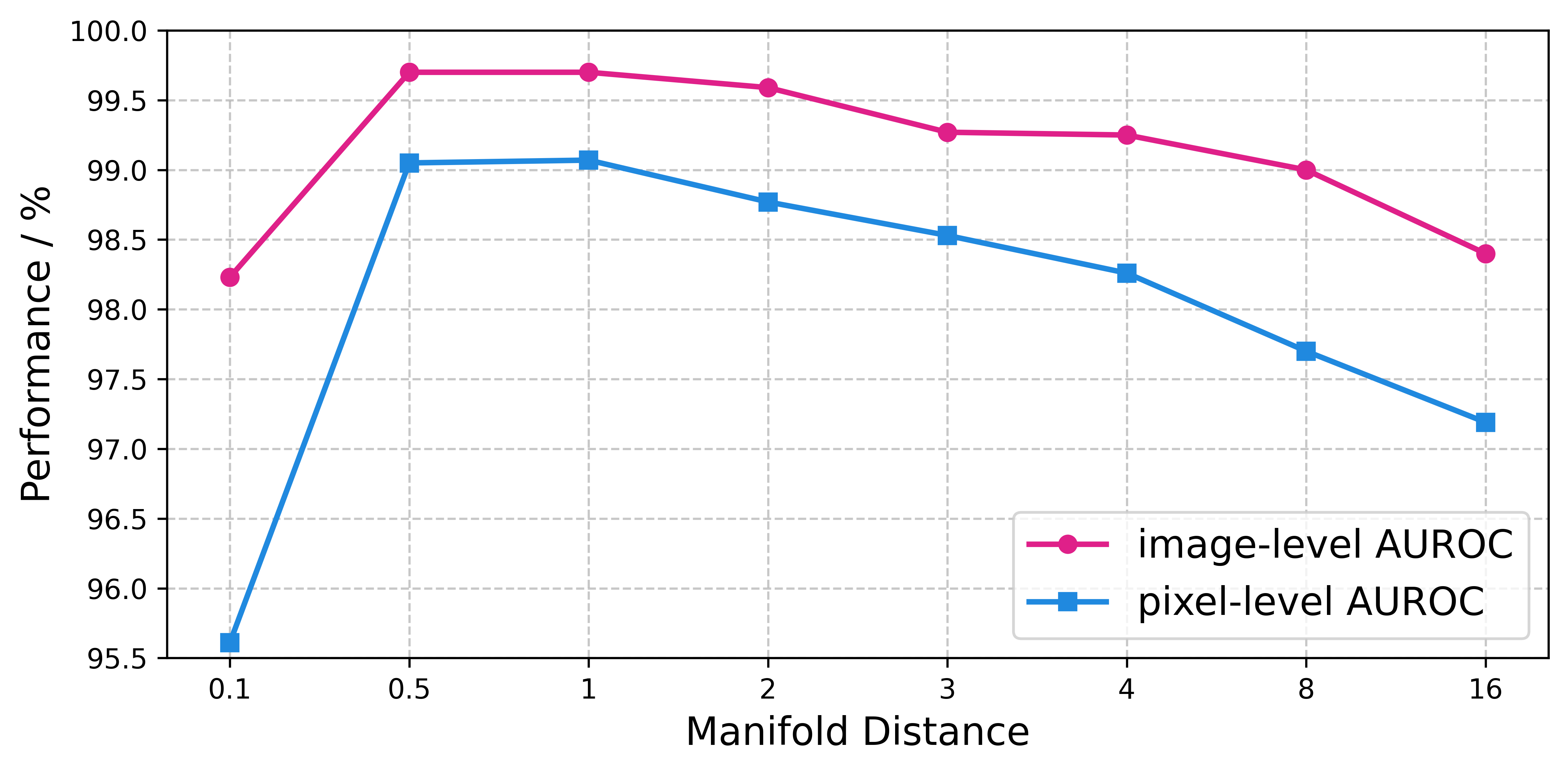}
    \caption{Performance of GLASS-m on MVTec AD under different manifold distance \(r_1\).}
    \label{fig:radius_selection}
  \end{minipage}\hfill
  \begin{minipage}{0.43\textwidth}
    \centering
    \captionof{table}{Performance of GLASS-m on MVTec AD under different backbone settings.
    ·/·/· same as above.}
    \resizebox{\textwidth}{!}{
      \begin{tabular}{l|c|c|c|c}
      \hline
      Backbone & level1 & level2 & level3 & MVTec AD \\
      \hline
      \multicolumn{1}{c|}{\multirow{7}[2]{*}{WideResNet50}} & \ding{51}     &       &       & 96.7/96.5/90.7 \\
            &       & \ding{51}     &       & 99.2/97.6/94.6 \\
            &       &       & \ding{51}     & \textbf{99.7}/98.9/94.6 \\
            &       & \ding{51}     & \ding{51}     & \textbf{99.7/99.1/96.4} \\
            & \ding{51}     & \ding{51}     &       & 98.4/97.5/92.9 \\
            & \ding{51}     &       & \ding{51}     & 99.0/98.3/92.3 \\
            & \ding{51}     & \ding{51}     & \ding{51}     & 99.2/98.6/94.1 \\
      \hline
      ResNet18 &       & \ding{51}     & \ding{51}     & 99.1/98.0/94.7 \\
      ResNet50 &       & \ding{51}     & \ding{51}     & 99.6/98.9/95.4 \\
      ResNet101 &       & \ding{51}     & \ding{51}     & 99.6/99.0/95.3 \\
      \hline
      \end{tabular}%
    }
    \label{tab:backbone_structure}
  \end{minipage}
\end{figure}

\noindent
\textbf{Dependency on Backbone.} As shown in Tab.~\ref{tab:backbone_structure}, merging the features output by level2 and level3 of
WideResNet50 achieves the best performance. We have chosen it as the default setting.
Meanwhile, our method does not depend on a specific backbone.
GLASS can maintain its good performance between several ResNets with different number of parameters on MVTec AD.

\noindent
\textbf{Feature Adaptor.} As introduced in Sec.~\ref{sec:_EandA}, we utilize the feature adapter \({{A}_{\varphi }}\) to mitigate latent domain bias brought by
the pre-trained backbone of feature extractor \( E_{\phi} \). We have conducted experiments using GLASS-m with and without \({{A}_{\varphi }}\) on MVTec AD.
As a result, the absence of \( A_{\varphi} \) in each branch leads to a decline of 0.1\% in pixel-level AUROC and 0.5\% in pixel-level PRO.

\noindent
\textbf{Manifold Distance.} We have introduced the manifold distance \(r_1\) for truncated projection of gradient ascent in Sec.~\ref{sec:_gas}.
It represents the relaxation tolerability for normal feature distribution which should not be too large (overfitting) or too small (underfitting),
facilitating controllable anomaly synthesis. As the pre-trained features have already been standardized,
the magnitude of gradient ascent distance \(\|\tilde{\varepsilon}_{i}^{h,w}\|\) mostly distributes around 1.
Fig.~\ref{fig:radius_selection} proves that the optimal range of \(r_1\) is \([0.5,1]\). Therefore, we have chosen \(r_1=1\) by default.

\section{Conclusion}
\label{sec:conclu}

In this paper, we propose a novel unified framework GLASS through the cooperative training of GAS and LAS for
synthesizing a broader coverage of anomalies in a controllable way under manifold and hypersphere hypothesis.
Specifically, we propose GAS based on gradient ascent and truncated projection. GAS has the capacity for quantitative synthesis of weak defects,
solving the problem of random synthetic direction in Gaussian noise.
LAS makes improvements by providing a more diverse range of anomaly synthesis. 
GLASS achieves SOTA results with faster detection speed on four anomaly detection datasets
in various industrial settings and shows superior performance in weak defect detection.
However, our main focus is localizing the structural anomalies in industrial scenarios.
We have not extensively explored the logical anomalies.
In the future, we will investigate the application of GLASS in logical anomaly detection
and plan to implement anomaly synthesis without relying on auxiliary texture datasets.

\section*{Acknowledgements}
\label{sec:acknow}
This work is supported by the National Natural Science Foundation of China under Grant 62303458, Grant 62303461 and Grant U21A20482.
This work is also supported by the Beijing Municipal Natural Science Foundation of China under Grant L243018.
In addition, we would like to express our gratitude to WEIQIAO Textile for collecting the original images used in the WFDD dataset.

\bibliographystyle{splncs04}

\begin{thebibliography}{10}
    \providecommand{\url}[1]{\texttt{#1}}
    \providecommand{\urlprefix}{URL }
    \providecommand{\doi}[1]{https://doi.org/#1}
    
    \bibitem{akcay2019ganomaly}
    Akcay, S., Atapour-Abarghouei, A., Breckon, T.P.: Ganomaly: Semi-supervised anomaly detection via adversarial training. In: Asia Conference on Computer Vision. pp. 622--637. Springer (2019). \doi{10.1007/978-3-030-20893-6_39}
    
    \bibitem{bae2023pni}
    Bae, J., Lee, J.H., Kim, S.: Pni: Industrial anomaly detection using position and neighborhood information. In: Proceedings of the IEEE/CVF International Conference on Computer Vision. pp. 6373--6383 (2023)
    
    \bibitem{batzner2024efficientad}
    Batzner, K., Heckler, L., K{\"o}nig, R.: Efficientad: Accurate visual anomaly detection at millisecond-level latencies. In: Proceedings of the IEEE/CVF Winter Conference on Applications of Computer Vision. pp. 128--138 (2024)
    
    \bibitem{bergmann2019mvtec}
    Bergmann, P., Fauser, M., Sattlegger, D., Steger, C.: Mvtec ad--a comprehensive real-world dataset for unsupervised anomaly detection. In: Proceedings of the IEEE/CVF Conference on Computer Vision and Pattern Recognition. pp. 9592--9600 (2019)
    
    \bibitem{cao2023collaborative}
    Cao, Y., Xu, X., Liu, Z., Shen, W.: Collaborative discrepancy optimization for reliable image anomaly localization. IEEE Transactions on Industrial Informatics  \textbf{19}(11),  10674--10683 (2023)
    
    \bibitem{cimpoi2014describing}
    Cimpoi, M., Maji, S., Kokkinos, I., Mohamed, S., Vedaldi, A.: Describing textures in the wild. In: Proceedings of the IEEE Conference on Computer Vision and Pattern Recognition. pp. 3606--3613 (2014)
    
    \bibitem{cubuk2020randaugment}
    Cubuk, E.D., Zoph, B., Shlens, J., Le, Q.V.: Randaugment: Practical automated data augmentation with a reduced search space. In: Proceedings of the IEEE/CVF Conference on Computer Vision and Pattern Recognition workshops. pp. 702--703 (2020)
    
    \bibitem{deng2022anomaly}
    Deng, H., Li, X.: Anomaly detection via reverse distillation from one-class embedding. In: Proceedings of the IEEE/CVF Conference on Computer Vision and Pattern Recognition. pp. 9737--9746 (2022)
    
    \bibitem{dinh2017density}
    Dinh, L., Sohl-Dickstein, J., Bengio, S.: Density estimation using real nvp. In: International Conference on Learning Representations (2017)
    
    \bibitem{goyal2020drocc}
    Goyal, S., Raghunathan, A., Jain, M., Simhadri, H.V., Jain, P.: Drocc: Deep robust one-class classification. In: International Conference on Machine Learning. vol.~119, pp. 3711--3721. PMLR (2020)
    
    \bibitem{gudovskiy2022cflow}
    Gudovskiy, D., Ishizaka, S., Kozuka, K.: Cflow-ad: Real-time unsupervised anomaly detection with localization via conditional normalizing flows. In: Proceedings of the IEEE/CVF Winter Conference on Applications of Computer Vision. pp. 98--107 (2022)
    
    \bibitem{hou2021divide}
    Hou, J., Zhang, Y., Zhong, Q., Xie, D., Pu, S., Zhou, H.: Divide-and-assemble: Learning block-wise memory for unsupervised anomaly detection. In: Proceedings of the IEEE/CVF International Conference on Computer Vision. pp. 8791--8800 (2021)
    
    \bibitem{hyun2024reconpatch}
    Hyun, J., Kim, S., Jeon, G., Kim, S.H., Bae, K., Kang, B.J.: Reconpatch: Contrastive patch representation learning for industrial anomaly detection. In: Proceedings of the IEEE/CVF Winter Conference on Applications of Computer Vision. pp. 2052--2061 (2024)
    
    \bibitem{jezek2021deep}
    Jezek, S., Jonak, M., Burget, R., Dvorak, P., Skotak, M.: Deep learning-based defect detection of metal parts: evaluating current methods in complex conditions. In: International congress on ultra modern telecommunications and control systems and workshops (ICUMT). pp. 66--71. IEEE (2021)
    
    \bibitem{lee2022cfa}
    Lee, S., Lee, S., Song, B.C.: Cfa: Coupled-hypersphere-based feature adaptation for target-oriented anomaly localization. IEEE Access  \textbf{10},  78446--78454 (2022)
    
    \bibitem{lei2023pyramidflow}
    Lei, J., Hu, X., Wang, Y., Liu, D.: Pyramidflow: High-resolution defect contrastive localization using pyramid normalizing flow. In: Proceedings of the IEEE/CVF Conference on Computer Vision and Pattern Recognition. pp. 14143--14152 (2023)
    
    \bibitem{li2021cutpaste}
    Li, C.L., Sohn, K., Yoon, J., Pfister, T.: Cutpaste: Self-supervised learning for anomaly detection and localization. In: Proceedings of the IEEE/CVF Conference on Computer Vision and Pattern Recognition. pp. 9664--9674 (2021)
    
    \bibitem{lin2017focal}
    Lin, T.Y., Goyal, P., Girshick, R., He, K., Doll{\'a}r, P.: Focal loss for dense object detection. In: Proceedings of the IEEE international conference on computer vision. pp. 2980--2988 (2017)
    
    \bibitem{liu2024deep}
    Liu, J., Xie, G., Wang, J., Li, S., Wang, C., Zheng, F., Jin, Y.: Deep industrial image anomaly detection: A survey. Machine Intelligence Research  \textbf{21}(1),  104--135 (2024). \doi{10.1007/s11633-023-1459-z}
    
    \bibitem{liu2023simplenet}
    Liu, Z., Zhou, Y., Xu, Y., Wang, Z.: Simplenet: A simple network for image anomaly detection and localization. In: Proceedings of the IEEE/CVF Conference on Computer Vision and Pattern Recognition. pp. 20402--20411 (2023)
    
    \bibitem{pless2009survey}
    Pless, R., Souvenir, R.: A survey of manifold learning for images. IPSJ Transactions on Computer Vision and Applications  \textbf{1},  83--94 (2009)
    
    \bibitem{reiss2021panda}
    Reiss, T., Cohen, N., Bergman, L., Hoshen, Y.: Panda: Adapting pretrained features for anomaly detection and segmentation. In: Proceedings of the IEEE/CVF Conference on Computer Vision and Pattern Recognition. pp. 2806--2814 (2021)
    
    \bibitem{roth2022towards}
    Roth, K., Pemula, L., Zepeda, J., Sch{\"o}lkopf, B., Brox, T., Gehler, P.: Towards total recall in industrial anomaly detection. In: Proceedings of the IEEE/CVF Conference on Computer Vision and Pattern Recognition. pp. 14318--14328 (2022)
    
    \bibitem{ruff2018deep}
    Ruff, L., Vandermeulen, R., Goernitz, N., Deecke, L., Siddiqui, S.A., Binder, A., M{\"u}ller, E., Kloft, M.: Deep one-class classification. In: International Conference on Machine Learning. pp. 4393--4402. PMLR (2018)
    
    \bibitem{salehi2021multiresolution}
    Salehi, M., Sadjadi, N., Baselizadeh, S., Rohban, M.H., Rabiee, H.R.: Multiresolution knowledge distillation for anomaly detection. In: Proceedings of the IEEE/CVF Conference on Computer Vision and Pattern Recognition. pp. 14902--14912 (2021)
    
    \bibitem{schluter2022natural}
    Schl{\"u}ter, H.M., Tan, J., Hou, B., Kainz, B.: Natural synthetic anomalies for self-supervised anomaly detection and localization. In: European Conference on Computer Vision. pp. 474--489. Springer (2022). \doi{10.1007/978-3-031-19821-2_27}
    
    \bibitem{scholkopf2001estimating}
    Sch{\"o}lkopf, B., Platt, J.C., Shawe-Taylor, J., Smola, A.J., Williamson, R.C.: Estimating the support of a high-dimensional distribution. Neural computation  \textbf{13}(7),  1443--1471 (2001)
    
    \bibitem{shrivastava2016training}
    Shrivastava, A., Gupta, A., Girshick, R.: Training region-based object detectors with online hard example mining. In: Proceedings of the IEEE Conference on Computer Vision and Pattern Recognition. pp. 761--769 (2016)
    
    \bibitem{tax2004support}
    Tax, D.M., Duin, R.P.: Support vector data description. Machine Learning  \textbf{54},  45--66 (2004)
    
    \bibitem{tien2023revisiting}
    Tien, T.D., Nguyen, A.T., Tran, N.H., Huy, T.D., Duong, S., Nguyen, C.D.T., Truong, S.Q.: Revisiting reverse distillation for anomaly detection. In: Proceedings of the IEEE/CVF Conference on Computer Vision and Pattern Recognition. pp. 24511--24520 (2023)
    
    \bibitem{xiao2023restricted}
    Xiao, F., Sun, R., Fan, J.: Restricted generative projection for one-class classification and anomaly detection. arXiv preprint arXiv:2307.04097  (2023)
    
    \bibitem{yang2023memseg}
    Yang, M., Wu, P., Feng, H.: Memseg: A semi-supervised method for image surface defect detection using differences and commonalities. Engineering Applications of Artificial Intelligence  \textbf{119},  105835 (2023)
    
    \bibitem{yao2023explicit}
    Yao, X., Li, R., Zhang, J., Sun, J., Zhang, C.: Explicit boundary guided semi-push-pull contrastive learning for supervised anomaly detection. In: Proceedings of the IEEE/CVF Conference on Computer Vision and Pattern Recognition. pp. 24490--24499 (2023)
    
    \bibitem{you2022unified}
    You, Z., Cui, L., Shen, Y., Yang, K., Lu, X., Zheng, Y., Le, X.: A unified model for multi-class anomaly detection. Advances in Neural Information Processing Systems  \textbf{35},  4571--4584 (2022)
    
    \bibitem{yu2021fastflow}
    Yu, J., Zheng, Y., Wang, X., Li, W., Wu, Y., Zhao, R., Wu, L.: Fastflow: Unsupervised anomaly detection and localization via 2d normalizing flows. arXiv preprint arXiv:2111.07677  (2021)
    
    \bibitem{zavrtanik2021draem}
    Zavrtanik, V., Kristan, M., Sko{\v{c}}aj, D.: Draem-a discriminatively trained reconstruction embedding for surface anomaly detection. In: Proceedings of the IEEE/CVF International Conference on Computer Vision. pp. 8330--8339 (2021)
    
    \bibitem{zavrtanik2021reconstruction}
    Zavrtanik, V., Kristan, M., Sko{\v{c}}aj, D.: Reconstruction by inpainting for visual anomaly detection. Pattern Recognition  \textbf{112},  107706 (2021)
    
    \bibitem{zavrtanik2022dsr}
    Zavrtanik, V., Kristan, M., Sko{\v{c}}aj, D.: Dsr--a dual subspace re-projection network for surface anomaly detection. In: European Conference on Computer Vision. pp. 539--554. Springer (2022). \doi{10.1007/978-3-031-19821-2_31}
    
    \bibitem{zhang2023destseg}
    Zhang, X., Li, S., Li, X., Huang, P., Shan, J., Chen, T.: Destseg: Segmentation guided denoising student-teacher for anomaly detection. In: Proceedings of the IEEE/CVF Conference on Computer Vision and Pattern Recognition. pp. 3914--3923 (2023)
    
    \bibitem{zhou2022rethinking}
    Zhou, Y.: Rethinking reconstruction autoencoder-based out-of-distribution detection. In: Proceedings of the IEEE/CVF Conference on Computer Vision and Pattern Recognition. pp. 7379--7387 (2022)
    
    \bibitem{zou2022spot}
    Zou, Y., Jeong, J., Pemula, L., Zhang, D., Dabeer, O.: Spot-the-difference self-supervised pre-training for anomaly detection and segmentation. In: European Conference on Computer Vision. pp. 392--408. Springer (2022). \doi{10.1007/978-3-031-20056-4_23}
    
\end{thebibliography}

\clearpage


\section*{Supplementary material}

# Appendix

## A. Dataset Details

Three public datasets and three self-built datasets are employed in the experiment. The numbers of images and classes in each dataset are detailed in Tab. S1. Three self-built datasets are released at github.com/cqylunlun/GLASS.

Table S1. Overview of all datasets employed in the experiment. N and A are abbreviations for 'normal' and 'anomaly', respectively.

| Dataset  | Train N. | Test N. | Test A. | Class Num. |
|----------|----------|---------|---------|------------|
| MVTec AD | 3629     | 467     | 1258    | 15         |
| VisA     | 8659     | 962     | 1200    | 12         |
| MPDD     | 888      | 176     | 282     | 6          |
| WFDD     | 3657     | 203     | 241     | 4          |
| MAD-man  | -        | 150     | 450     | 15         |
| MAD-sys  | -        | 1600    | 4730    | 5          |

### A.1. Public Datasets

**MVTec AD.** The MVTec Anomaly Detection [4] dataset contains 15 categories of high-resolution industrial products with 5354 images, including over 70 types of defects.

**VisA.** The Visual Anomaly [41] dataset is the largest industrial anomaly detection dataset, including 10821 images across 12 categories of colored industrial parts.

**MPDD.** The Metal Parts Defect Detection [14] dataset contains 1346 images of metal parts under varied camera conditions across 6 categories.

### A.2. Self-built Datasets

**WFDD.** To further demonstrate the performance of GLASS in real-world industrial scenarios, we construct a Woven Fabric Defect Detection (WFDD) dataset, comprising 4101 images of woven fabrics with different textures and patterns across 4 categories. Specifically, the training set includes 3657 normal samples, while the test set consists of 203 normal samples and 241 anomaly samples. As shown in Fig. S1, WFDD comprises 3 categories of regularly textured samples collected from the industrial production sites of WEIQIAO Textile and 1 category of irregular patterned samples gathered from the publicly available Cloth Flaw Dataset[1]. Apart from the normal samples, each category contains block-shape, point-like, and line-type defects with pixel-level annotations.

**MAD-man.** To evaluate the detection ability of GLASS on real weak defects, we construct a test set named MVTec AD-manual (MAD-man) obtained by 5 individuals who independently select weak defect samples from all 15 categories of MVTec AD under unbiased conditions. Subjective interference is eliminated through the selection by multiple individuals. Each category of the 5 subsets contains 2

[1] tianchi.aliyun.com/dataset/79336?lang=en-us

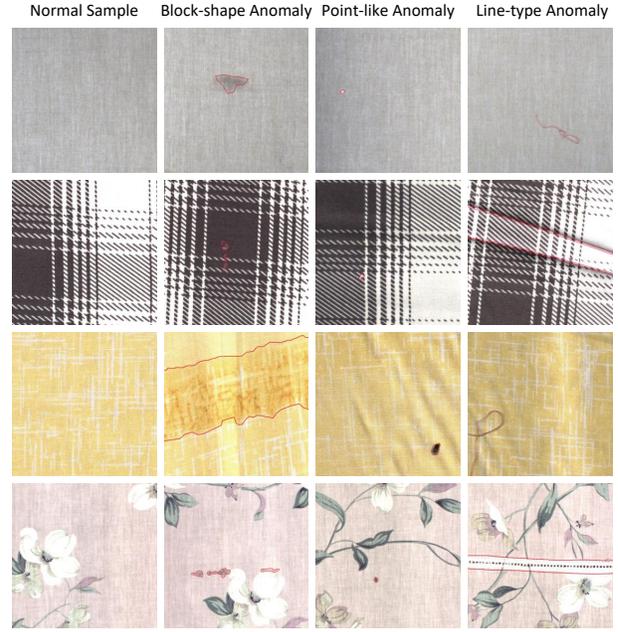

Figure S1. Examples of the self-built dataset WFDD. From top to bottom, the four categories are grey cloth, grid cloth, yellow cloth, and pink flower, respectively. The pixel-level annotations are marked by the red boundary on anomaly samples.

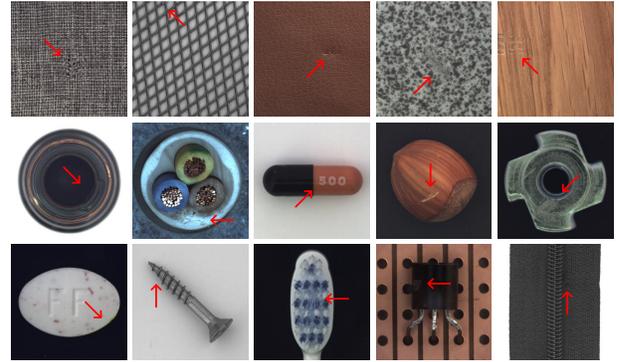

Figure S2. Selected weak defect examples from self-built test set MAD-man. The red arrow indicates the location of anomaly.

normal and 6 anomaly samples. Fig. S2 displays the weak defects selected by the first person across all categories.

**MAD-sys.** Due to the scarcity of weak defects in MVTec AD, we construct a test set named MVTec AD-synthesis (MAD-sys) from 5 texture categories of it. By adjusting the values of transparency coefficient $\beta = \{0.1, 0.3, 0.5, 0.7\}$ in Eq. 4, MAD-sys contains 4 subsets with varying degrees of weak defects. In each anomaly image of MAD-sys, the normal background is from the train set and the anomalous foreground is obtained by augmenting another random train image. Fig. S3 shows that the weak defects gradually become more difficult to distinguish with the increase of $\beta$.

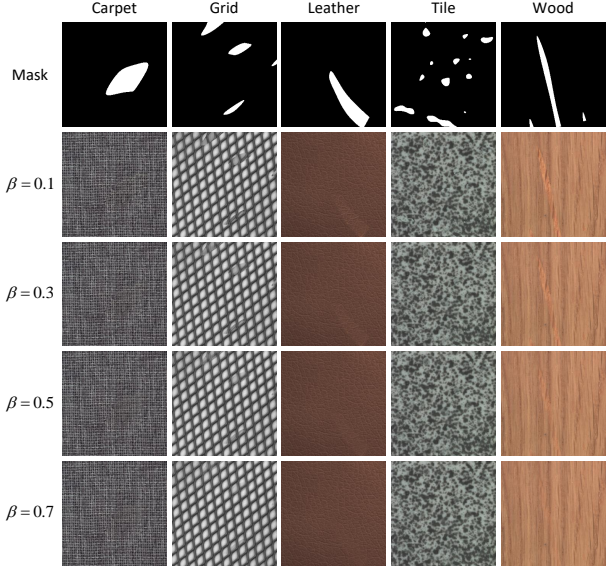

Figure S3. Synthetic weak defect examples from self-built test set MAD-sys under different transparency coefficient $\beta$.

## B. Implementation Details

### B.1. GAS under Hypersphere Hypothesis

The manifold and hypersphere hypotheses (in Sec. 3.2) are quite similar. Due to the reinforcement of the connection between global and local anomaly features, the truncated projection process in hypersphere hypothesis differs from that in manifold. Therefore, GAS algorithm under hypersphere hypothesis is provided in Alg. S1.

### B.2. Image-level Spectrogram Analysis

GLASS-m and GLASS-h (in Sec. 3.2) are based on manifold and hypersphere hypothesis, respectively. GLASS-j integrates GLASS-h and GLASS-m through image-level spectrogram analysis. In practical applications, image-level spectrogram is analyzed to determine the distribution hypothesis of different categories. Specifically, we compute the average of all samples $\bar{x} = \frac{1}{|X_{\text{train}}|} \sum_{x_i \in X_{\text{train}}} x_i$ for each category. Subsequently, the Discrete Fourier Transform (DFT) is applied to obtain the spectrogram $S_i = f_{DFT}(x_i)$. The compactness of $S_i$ through central shifting and binarization is calculated by the proportion of positive pixels within the area enclosed by the orange boundary in Fig. S5.

By setting a general threshold, we can determine the distribution hypothesis for each category. Lesser compactness tends to manifold hypothesis, whereas greater compactness tends to hypersphere hypothesis. As illustrated in Fig. S5, category of hypersphere distribution shows more concentrated intraclass distribution, while category of manifold distribution shows more complex nonlinear structures in each sample. Comparative experiments in Sec. D indicate that most hypothesis derived from image-level spectrogram analysis are more effective than the opposite hypothesis.

## C. Extended Ablation Study

### C.1. Neighborhood Size

As mentioned in Sec. 3.1, the neighborhood size $p$ is the patch size of feature aggregation in feature extractor $E_\phi$. Fig. S4 indicates that the performance difference among $p = 3$, $p = 4$, and $p = 5$ is minor. We choose $p = 3$ yielding the highest pixel-level AUROC.

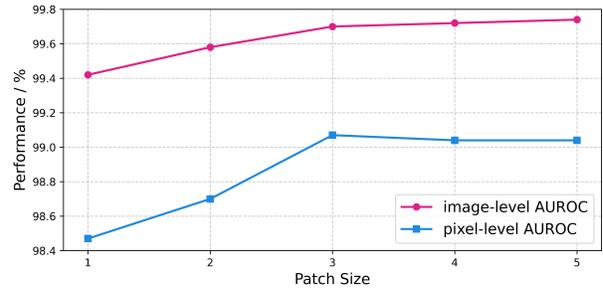

Figure S4. Performance of GLASS-m on MVTec AD under different patch sizes.

---

**Algorithm S1** GAS under Hypersphere Hypothesis
---
1: **Input:** normal feature map $u_i$, center of normal features $c$, LAS feature map $u_{i+}$, number of batch $n_{\text{batch}}$, number of iteration $n_{\text{step}}$, interval of projection $n_{\text{proj}}$
2: **Output:** global anomaly feature maps $\tilde{v}_i$
3: **for** batch $= 1$ to $n_{\text{batch}}$ **do**
4:     Initialize $u_i$ by $E_\phi$ and $A_\varphi$
5:     *Gaussian noise*. Add $\varepsilon_i$ to $u_i \to g_i$
6:     **for** step $= 1$ to $n_{\text{step}}$ **do**
7:         *Gradient ascent*.
8:         (a) Calculate GAS branch loss $L_{\text{gas}}$ by Eq. 6
9:         (b) Update $\tilde{g}_i$ according to Eq. 1 with no grad.
10:        **if** step is a multiple of $n_{\text{proj}}$ **then**
11:           *Truncated projection* $r'_1 \sim r'_2$.
12:           (c) Get gradient ascent dist. $\tilde{\varepsilon}_i = \tilde{g}_i - c$
13:           (d) Constrain range like Eq. 2 to get $\tilde{\varepsilon}_i \to \hat{\varepsilon}_i$
14:           (e) Get GAS feature $\tilde{v}_i = c + \hat{\varepsilon}_i$
15:        **end if**
16:     **end for**
17:     *Truncated projection* $r'_2 \sim r'_3$.
18:     (f) Get gradient ascent dist. $\tilde{\varepsilon}'_i = u_{i+} - c$
19:     (g) Constrain range like Eq. 2 to get $\tilde{\varepsilon}'_i \to \hat{\varepsilon}'_i$
20:     (h) Get LAS feature $u'_{i+} = c + \hat{\varepsilon}'_i$
21: **end for**
22: **return** $\tilde{v}_i$

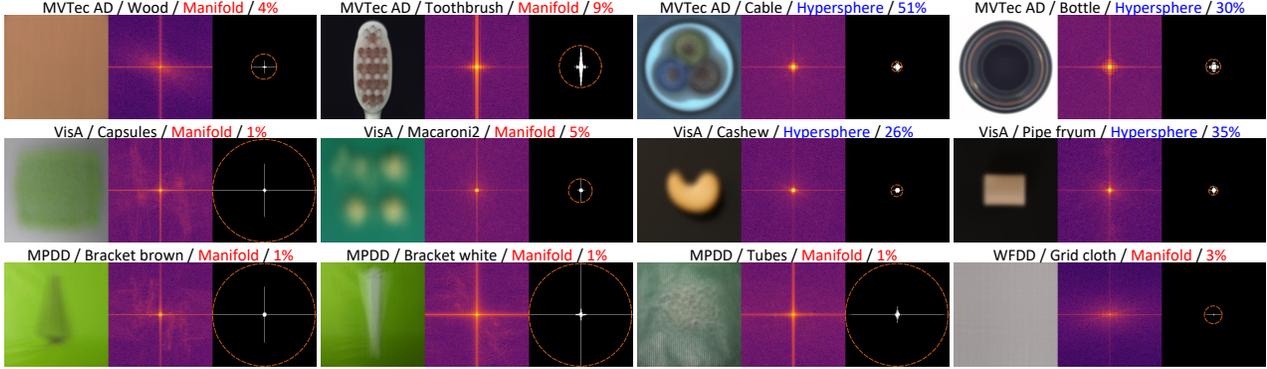

Figure S5. Examples of image-level spectrogram analysis on four datasets. Each figure contains three subfigures, representing the average of all samples, grayscale spectrogram, and binary spectrogram, from left to right. The compactness is calculated by the proportion of positive pixels within the orange boundary area. ·/·/·/· denotes dataset name, class name, inferred hypothesis and spectrogram compactness.

## C.2. Scale of Noise

As discussed in Sec. 3.2, Gaussian noise is utilized to synthesize anomalies in feature space. However, due to the constraint of gradient ascent and truncated projection, the scale of Gaussian noise has a wide range of options. Fig. S6 shows that the performance gap under different standard deviations is negligible. Similarly, we choose $\sigma_g = 0.015$ yielding the highest pixel-level AUROC.

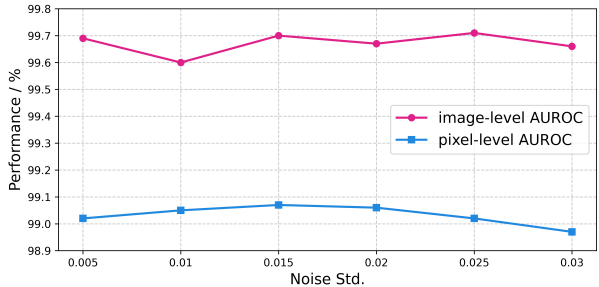

Figure S6. Performance of GLASS-m on MVTec AD under different scales of noise.

## C.3. Components in LAS

As proposed in Sec. 3.3, LAS utilizes three operations of interaction, union, and single to obtain anomaly masks. Tab. S2 indicates that the utilization of three operations, which exhibit the greatest diversity in anomalous regions, outperforms the others. Furthermore, both of the ambiguous synthesis position in the absence of the foreground mask and the reduced anomaly diversity with the constant $\beta$ will lead to a decline in performance slightly. Due to the superior synthesis diversity, LAS outperforms CutPaste [17] and NSA [26]. Our GLASS ("GAS + LAS") improves the average I-AUROC, P-AUROC, and P-PRO by margins of 0.5%, 0.6%, and 3.2% compared to "GAS + CutPaste",

Table S2. Performance of GLASS-m on MVTec AD under different anomaly synthesis strategies with the proposed GAS. ·/·/· denotes image-level AUROC%, pixel-level AUROC%, and pixel-level PRO%. N/C is the abbreviation of no change.

| LAS | | | | MVTec AD |
|---|---|---|---|---|
| Interaction | Union | Single | Others | |
| ✓ | | | N/C | 98.2/98.3/94.9 |
| | ✓ | | N/C | **99.7**/99.0/96.2 |
| ✓ | | ✓ | N/C | **99.7**/99.0/96.3 |
| | ✓ | ✓ | N/C | **99.7**/99.0/96.3 |
| ✓ | ✓ | ✓ | N/C | **99.7/99.1/96.4** |
| ✓ | ✓ | ✓ | w/ $\beta = 0.5$ | **99.7**/99.0/95.9 |
| ✓ | ✓ | ✓ | w/o fgmask | 99.6/99.0/95.8 |

Table S3. Performance of GLASS-m on MVTec AD under different training objectives for the three branches. ·/·/· denotes image-level AUROC%, pixel-level AUROC%, and pixel-level PRO%.

| Normal | GAS | LAS | MVTec AD |
|---|---|---|---|
| BCE | BCE | BCE | 99.4/98.9/93.9 |
| BCE | BCE | Focal | 99.6/99.0/95.3 |
| BCE | BCE | Focal+OHEM | **99.7/99.1/96.4** |

and by 0.2%, 0.4%, and 3.2% compared to "GAS + NSA".

## C.4. Training Objectives

As introduced in Sec. 3.4, the commonly used BCE loss is utilized for the training phase of the Normal and GAS branches. However, the detection of local anomaly features synthesized by LAS can be viewed as a segmentation task, facing the issue of imbalance between positive and negative samples. Therefore, the discriminator $D_\psi$ in LAS branch is trained by the Focal loss. Tab. S3 demonstrates that using Focal loss for LAS can increase the pixel-level PRO by 1.4%. Additionally, by using OHEM to filter crucial samples, PRO is further increased by 1.1%. Despite already achieving good performance with BCE loss, the introduction of Focal loss and OHEM brings better performance.

Table S4. Comparison of GLASS and its variants with different SOTA methods on each category of VisA. ·/· denotes image-level AUROC% and pixel-level AUROC%.

| Category | DSR | PatchCore | BGAD | RD++ | SimpleNet | GLASS-m | GLASS-h | GLASS-j |
| --- | --- | --- | --- | --- | --- | --- | --- | --- |
| Candle | 80.4/72.1 | **99.5/99.3** | 97.7/98.9 | 96.0/98.8 | 98.5/98.8 | 99.3/99.2 | 98.9/99.4 | 99.3/99.2 |
| Capsules | 87.4/94.4 | 76.8/**99.3** | 91.2/99.0 | 92.4/99.4 | 92.4/98.8 | **96.8/99.3** | 91.0/97.3 | **96.8/99.3** |
| Cashew | 90.8/97.9 | 97.8/98.6 | 96.1/97.7 | 97.8/95.5 | 97.3/**98.9** | 98.3/98.5 | **98.9/98.9** | **98.9/98.9** |
| Chewinggum | 95.7/96.0 | 99.1/98.9 | 99.6/98.8 | 99.1/98.4 | 99.9/98.5 | **100/99.4** | 99.8/99.1 | **100/99.4** |
| Fryum | 92.5/50.4 | 96.0/92.6 | 98.0/**96.8** | 97.0/96.6 | 98.7/90.0 | 98.7/94.9 | **99.0**/94.8 | **99.0**/94.8 |
| Macaroni1 | 89.6/78.8 | 97.6/99.7 | 97.9/99.3 | 94.1/99.7 | 99.5/99.6 | **100/99.8** | 99.8/99.7 | **100/99.8** |
| Macaroni2 | 74.4/86.2 | 76.0/98.7 | 91.3/99.2 | 88.6/**99.7** | 80.8/99.0 | **96.2/99.7** | 95.1/99.6 | **96.2/99.7** |
| Pcb1 | 79.8/78.7 | 98.6/**99.8** | 95.6/99.0 | 96.8/99.7 | **99.8**/99.7 | 98.5/99.3 | 98.7/99.4 | 98.7/99.4 |
| Pcb2 | 93.8/85.3 | 97.2/98.8 | 94.9/98.5 | 96.7/**98.9** | **99.5**/98.2 | 98.5/97.9 | 98.4/98.3 | 98.4/98.3 |
| Pcb3 | 97.2/87.8 | 98.7/**99.4** | 96.5/98.7 | 97.3/99.3 | 99.2/99.2 | **99.5**/99.3 | 99.3/99.0 | **99.5**/99.3 |
| Pcb4 | 97.9/96.7 | **99.8**/98.2 | 98.3/98.0 | 99.7/**98.8** | 99.4/98.3 | 99.3/98.8 | 99.1/98.6 | 99.1/98.6 |
| Pipe fryum | 76.8/87.6 | 99.8/98.9 | 99.3/98.9 | 99.8/**99.1** | 99.9/99.0 | **100**/98.9 | **100**/99.1 | **100**/99.1 |
| Average | 88.0/84.3 | 94.7/98.5 | 96.4/98.6 | 96.3/98.7 | 97.1/98.2 | **98.8**/98.7 | 98.2/98.6 | **98.8/98.8** |

Table S5. Comparison of GLASS and its variants with different SOTA methods on each category of MPDD. ·/· denotes image-level AUROC% and pixel-level AUROC%.

| Category | DSR | PatchCore | BGAD | RD++ | SimpleNet | GLASS-m | GLASS-h | GLASS-j |
| --- | --- | --- | --- | --- | --- | --- | --- | --- |
| Bracket black | 44.6/76.2 | 86.8/98.1 | 82.9/97.9 | 90.5/98.2 | 94.2/98.3 | **99.7/99.6** | 95.2/99.2 | **99.7/99.6** |
| Bracket brown | 66.7/40.4 | 95.1/98.3 | 98.4/97.3 | 95.0/97.2 | 96.0/96.6 | **98.6/98.4** | 91.6/96.6 | **98.6/98.4** |
| Bracket white | 77.6/94.9 | 88.6/**99.8** | 77.9/98.3 | 91.7/99.4 | 98.4/99.6 | **99.7**/99.2 | 94.4/98.7 | **99.7**/99.2 |
| connector | **100**/82.8 | **100**/99.5 | 96.4/98.6 | **100**/99.4 | **100**/99.5 | **100/99.8** | **100**/99.7 | **100/99.8** |
| Metal plate | **100**/69.0 | **100**/98.6 | 96.5/97.4 | **100**/99.1 | **100**/98.5 | **100/99.5** | **100**/99.4 | **100/99.5** |
| Tubes | 97.2/94.0 | 90.8/98.8 | 98.7/99.4 | 95.7/99.1 | **99.7**/99.5 | 99.4/**99.7** | 99.2/99.4 | 99.4/**99.7** |
| Average | 81.0/76.2 | 93.5/98.9 | 91.8/98.1 | 95.5/98.7 | 98.1/98.7 | **99.6/99.4** | 96.7/98.8 | **99.6/99.4** |

Table S6. Comparison of GLASS and its variants with different SOTA methods on each category of WFDD. ·/· denotes image-level AUROC% and pixel-level AUROC%.

| Category | DSR | PatchCore | BGAD | RD++ | SimpleNet | GLASS-m | GLASS-h | GLASS-j |
| --- | --- | --- | --- | --- | --- | --- | --- | --- |
| Grey cloth | **100**/95.7 | **100**/98.2 | **100**/98.8 | 99.4/98.2 | 99.8/97.5 | **100**/99.2 | **100**/99.2 | **100**/99.2 |
| Grid cloth | 99.0/94.9 | **100**/97.9 | **100**/98.1 | 97.5/98.1 | 99.5/97.4 | 99.9/**98.5** | 97.8/97.3 | 99.9/**98.5** |
| Yellow cloth | 99.9/80.3 | **100**/96.6 | **100**/97.6 | **100**/97.6 | 99.9/97.1 | **100**/98.2 | 99.8/97.3 | **100**/98.2 |
| Pink flower | 81.4/80.6 | 85.3/99.8 | 88.5/99.5 | 84.1/99.7 | 96.0/99.8 | **100/99.9** | 98.4/99.7 | **100/99.9** |
| Average | 95.1/87.9 | 96.3/98.1 | 97.1/98.5 | 95.2/98.4 | 98.8/98.0 | **100/98.9** | 99.0/98.4 | **100/98.9** |

Table S7. Comparison of GLASS-m with different SOTA methods on MAD-man. ·/·/· denotes image-level AUROC%, pixel-level AUROC%, and pixel-level PRO%.

| MAD-man | DSR | RD++ | PatchCore | BGAD | SimpleNet | GLASS |
| --- | --- | --- | --- | --- | --- | --- |
| Person 1 | 93.3/96.3/89.7 | **100**/98.9/96.8 | 98.9/98.5/92.7 | 97.8/98.5/95.3 | **100**/98.5/94.1 | **100/99.2/97.4** |
| Person 2 | 93.3/97.9/93.7 | 98.9/99.2/97.1 | 98.3/98.9/96.5 | 98.9/99.2/97.5 | **99.4**/98.9/95.9 | **99.4/99.6/98.4** |
| Person 3 | 95.6/98.4/95.0 | 97.8/99.3/97.6 | 97.2/99.3/97.1 | 94.4/99.1/97.3 | 99.4/99.2/97.4 | **99.4/99.6/98.4** |
| Person 4 | 93.3/93.8/85.6 | 96.1/98.6/94.8 | 94.4/98.1/91.3 | 92.8/98.0/93.3 | 98.3/98.1/91.5 | **98.9/99.2/96.3** |
| Person 5 | 95.6/98.1/92.5 | 98.3/98.8/97.0 | 98.9/98.2/95.8 | 95.6/98.7/96.9 | 99.4/98.4/94.9 | **100/99.2/97.4** |
| Average | 94.2/96.9/91.3 | 98.2/98.9/96.7 | 97.6/98.6/94.7 | 95.9/98.7/96.1 | 99.3/98.6/94.8 | **99.6/99.3/97.5** |

Table S8. Comparison of GLASS-m with different SOTA methods on MAD-sys. ·/·/· denotes image-level AUROC%, pixel-level AUROC%, and pixel-level PRO%.

| MAD-sys | DSR | PatchCore | BGAD | RD++ | SimpleNet | GLASS |
| --- | --- | --- | --- | --- | --- | --- |
| $\beta = 0.1$ | 97.8/96.2/90.2 | 97.9/96.4/75.0 | 97.3/93.1/77.6 | 91.2/90.8/71.9 | 92.7/90.2/69.0 | **99.4/98.2/92.0** |
| $\beta = 0.3$ | 95.0/92.8/84.4 | 96.8/96.8/71.5 | 95.6/95.6/71.6 | 90.2/90.2/72.1 | 91.1/91.1/61.5 | **98.4/98.4/85.5** |
| $\beta = 0.5$ | 90.6/88.3/75.6 | 94.1/94.1/63.6 | 91.4/87.3/62.2 | 85.2/87.7/60.5 | 86.2/86.6/57.5 | **96.8/94.4/80.7** |
| $\beta = 0.7$ | 82.6/83.3/62.9 | 80.7/**83.7**/45.4 | 77.4/75.4/44.6 | 69.0/76.3/42.5 | 68.5/73.2/41.0 | **88.0**/83.7/**63.1** |
| Average | 91.5/90.2/78.3 | 92.4/92.6/63.9 | 90.4/86.8/64.0 | 83.9/86.3/61.8 | 84.6/85.0/57.3 | **95.6/93.3/80.3** |

## D. Detailed Comparative Experiments

According to [19], five SOTA methods are employed, including DSR [38] based on image reconstruction, PatchCore [23] based on memory bank, BGAD$^{w/o}$ [33] based on normalizing flow, RD++ [30] based on knowledge distillation, and SimpleNet [20] based on one-class classification. Specifically, DSR and SimpleNet utilize feature-level anomaly synthesis, while BGAD and RD++ utilize image-level anomaly synthesis.

### D.1. Anomaly Detection on Public Datasets

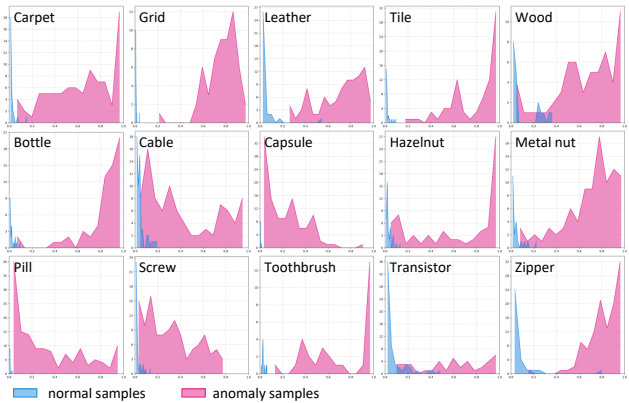

Figure S7. Histogram of anomaly scores of DSR on MVTec AD.

**Anomaly Detection on MVTec AD.** The superior performance of GLASS on MVTec AD is discussed in Sec. 4.3. As shown in Tab. 1, GLASS-j achieves SOTA performance with an average of **99.9%** image-level AUROC and **99.3%** pixel-level AUROC. As demonstrated in Fig. 5 and Fig. S7, due to lesser overlapping areas and more concentrated anomaly scores, GLASS demonstrates better discriminative ability than DSR in distinguishing between normal samples and anomaly samples.

**Anomaly Detection on VisA.** GLASS-j achieves superior performance on 8 out of 12 categories as shown in Tab. S4, establishing itself as the SOTA method for VisA with an average of **98.8%** image-level AUROC and **98.8%** pixel-level AUROC. Moreover, all three variants of GLASS outperform other SOTA methods. Due to the complex nonlinear structure of most samples, the locally-sensitive GLASS-m performs slightly better than GLASS-h.

**Anomaly Detection on MPDD.** GLASS-j achieves superior performance on all categories as shown in Tab. S5, establishing itself as the SOTA method for MPDD with an average of **99.6%** image-level AUROC and **99.4%** pixel-level AUROC. Since all categories are judged to be manifold distribution, GLASS-j is equal to GLASS-m in the anomaly detection task of MPDD. As a result of the extremely low spectrogram compactness of 'Bracket brown' and 'Bracket white' classes in Fig. S5, GLASS-m significantly outperforms GLASS-h on these two categories.

**Anomaly Detection under complex environments.** As shown in Fig. S8, several categories from three public datasets already include samples with various inclination angles and scattered objects. Due to the lack of samples with dirty backgrounds, we simulate such conditions on the 'Capsules' class. We keep the foreground of samples unchanged, while augmenting the background by applying color jittering and adding different scales of Gaussian noise. Compared to the clean backgrounds, GLASS shows a decrease of 0.7% in I-AUROC on the dirty backgrounds, while SimpleNet [20] shows a larger decrease of 6.2%. Therefore, GLASS can handle samples with complex industrial environments, such as scattered objects and dirty backgrounds.

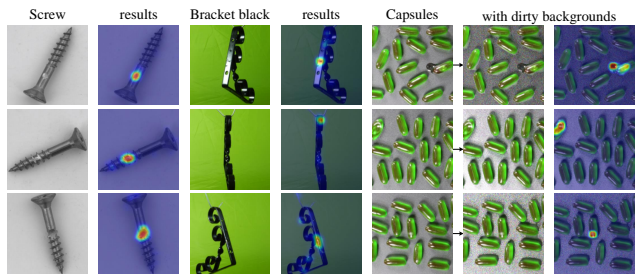

Figure S8. Test anomalies from MVTec AD, MPDD, and VisA.

### D.2. Anomaly Detection on Self-built Datasets

**Anomaly Detection on WFDD.** GLASS-j achieves superior performance on all categories as shown in Tab. S6 with an average of **100%** image-level AUROC and **98.9%** pixel-level AUROC on WFDD. Similarly, GLASS-j is equal to GLASS-m in the anomaly detection task of WFDD. It is evident that GLASS-m demonstrates significantly better performance in texture categories compared to GLASS-h, indicating that GLASS-h is more inclined towards categories with more concentrated intraclass distribution.

**Anomaly Detection on MAD-man.** MAD-man is individually selected by five people under unbiased conditions. As shown in Tab. S7, GLASS achieves superior performance on MAD-man with the selected weak defects. Meanwhile, GLASS outperforms other methods in all three metrics for each subset. Consequently, GLASS has a greater advantage in weak defect detection.

**Anomaly Detection on MAD-sys.** As shown in Tab. S8, GLASS achieves superior performance on MAD-sys with the synthetic weak defects. GLASS significantly outperforms other SOTA methods in detecting different degree of weak defects. Additionally, as $\beta$ increases and the defect regions become less distinguishable, the advantage of GLASS over others becomes more pronounced with the gradual decline in performance.

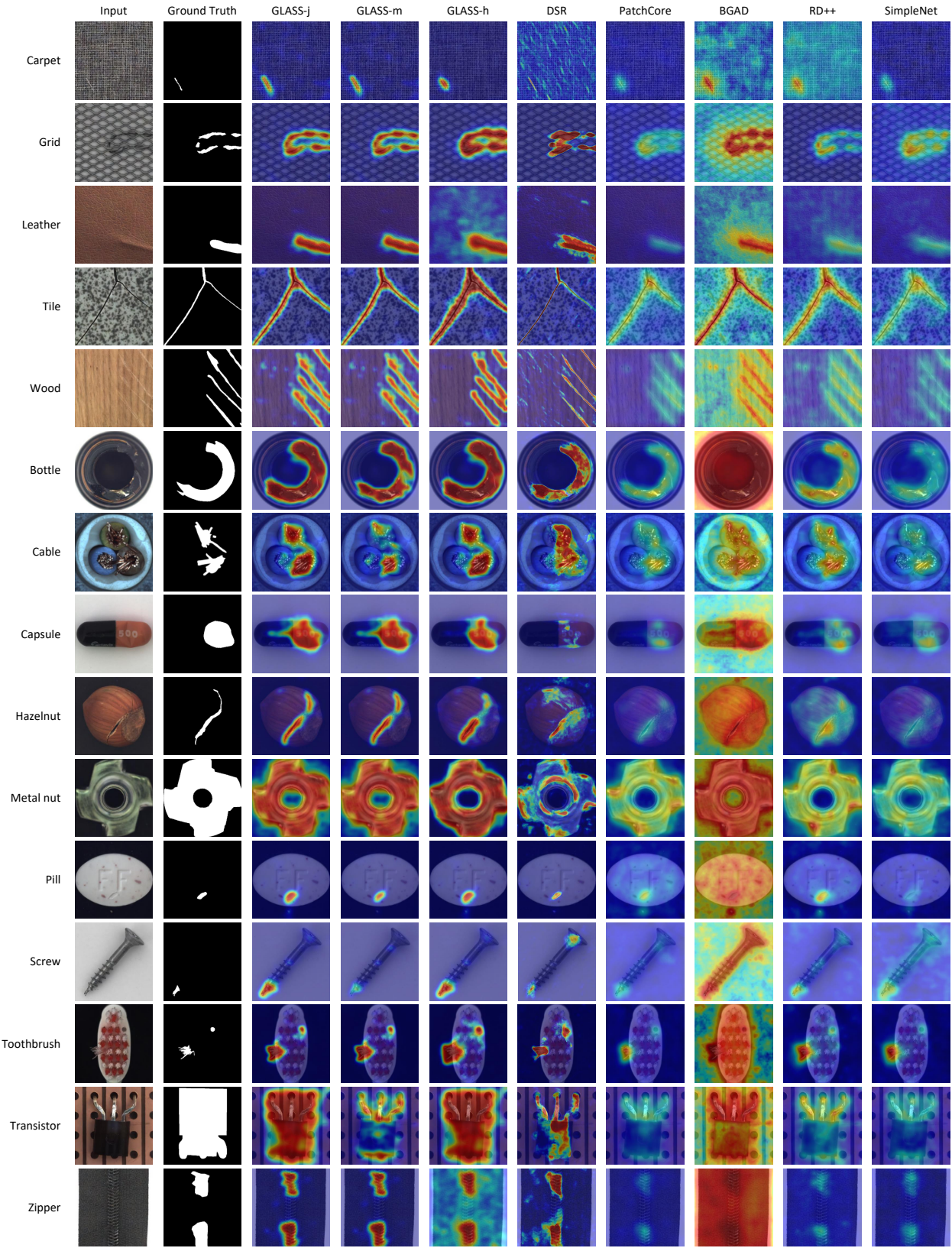

Figure S9. Visualizations of GLASS and its variants compared with different SOTA methods on MVTec AD.

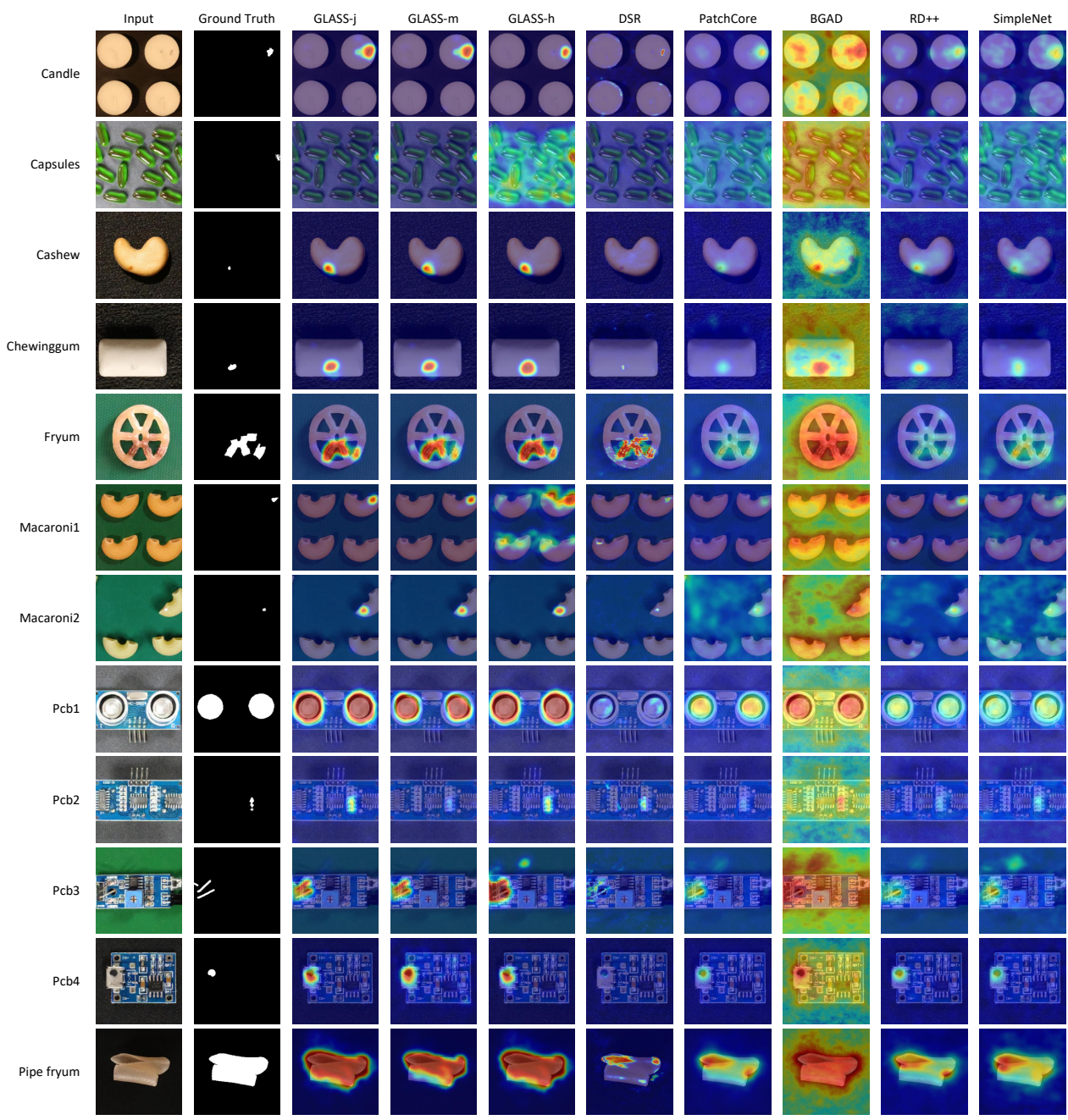

Figure S10. Visualizations of GLASS and its variants compared with different SOTA methods on VisA.

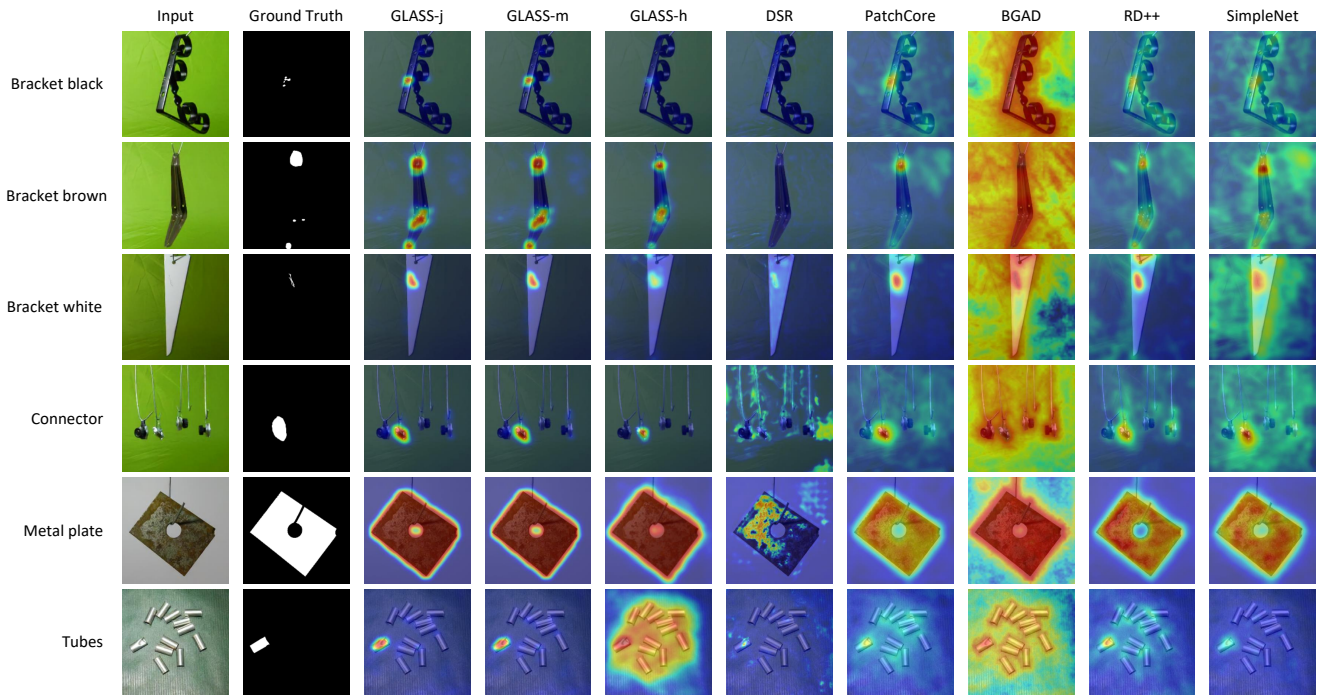

Figure S11. Visualizations of GLASS and its variants compared with different SOTA methods on MPDD.

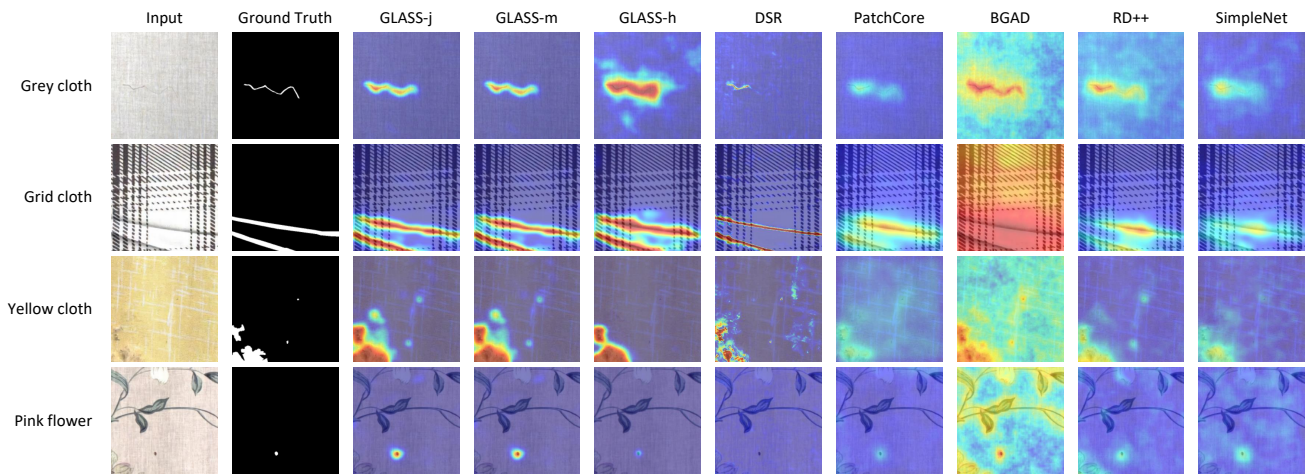

Figure S12. Visualizations of GLASS and its variants compared with different SOTA methods on WFDD.

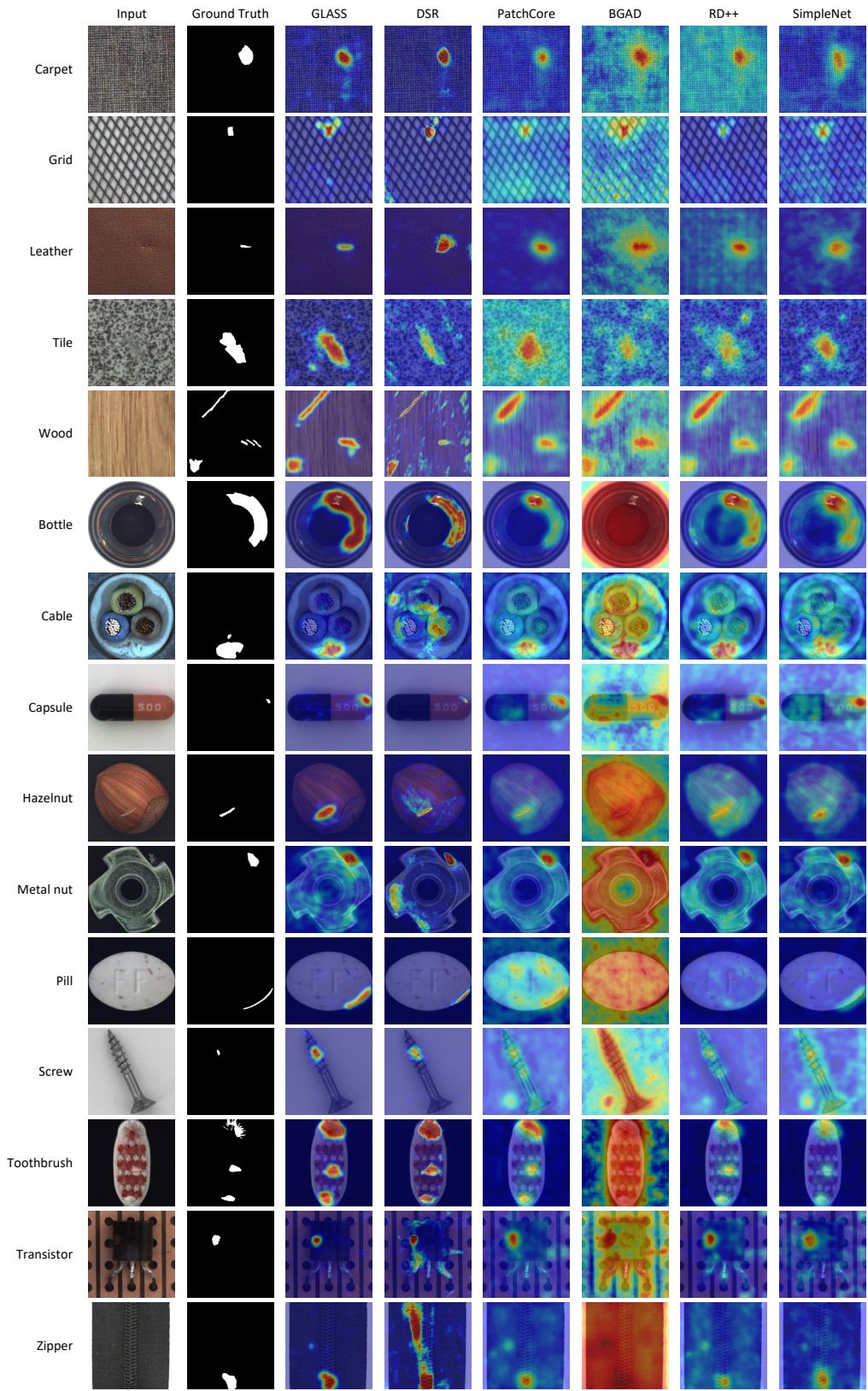

Figure S13. Visualizations of GLASS-m compared with different SOTA methods on selected weak defect test set MAD-man.

# E. Qualitative Results

We present the visualization results of three GLASS variants and five comparative methods across six datasets. To clearly show the relative confidence outputs in different areas, anomaly maps of all samples in each category are processed by normalization at unified scales, rather than individual normalization at varying scales for each sample.

## E.1. Visualizations on Public Datasets

**Visualizations on MVTec AD.** As shown in Fig. S9, GLASS ensures the lower anomaly score of normal areas, while accurately localizing anomalous regions with higher confidence for both texture and object categories. Specifically, results on the 'Carpet' and 'Pill' classes demonstrate that GLASS is less prone to over-detection and missed detection. Furthermore, results on the 'Metal nut' and 'Transistor' classes reveal the superiority of GLASS to detect global anomalies.

**Visualizations on VisA.** As shown in Fig. S10, GLASS demonstrates better detection performance in areas with relatively small anomalies. Moreover, results on the 'Capsules' and 'Macaroni1' classes indicate that GLASS can precisely locate anomalies even when multiple objects of the same type exist in the image.

**Visualizations on MPDD.** Compared to other SOTA methods, the results of GLASS have clearer edges, more complete structures, and better consistency as shown in Fig. S11. Since GLASS-j determines the distribution hypothesis through judgment, it consistently aligns with either GLASS-m or GLASS-h in each category. Due to the dispersed intraclass distribution in MPDD, GLASS-m yields better anomaly localization performance than GLASS-h.

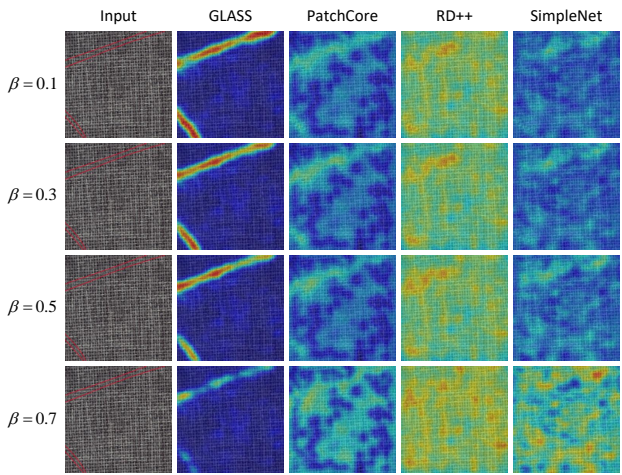

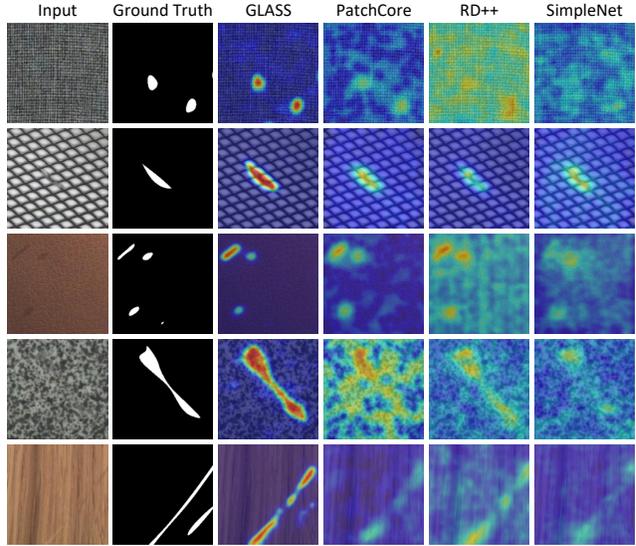

Figure S15. Visualizations of GLASS-m compared with different SOTA methods on synthetic weak defect test set MAD-sys.

## E.2. Visualizations on Self-built Datasets

**Visualizations on WFDD.** Fig. S12 shows that GLASS can effectively detect continuous line-type defects. In addition, the performance of GLASS is not significantly affected by complex patterned backgrounds, as evidenced by the results on the 'Pink flower' class. Furthermore, results on the 'Yellow cloth' class reveal that GLASS has the capability to distinguish extremely small anomalous areas, such as those less than $10 \times 10$ pixels.

**Visualizations on MAD-man.** Fig. S13 demonstrates the detection results of the weak defects selected from all 15 categories of MVTec AD. Although most defects are weak and difficult to distinguish, GLASS still ensures low confidence levels in normal areas and high confidence levels in anomalous regions. Specifically, results on the 'Bottle' class show that, unlike other methods that focus more on obvious anomalies, GLASS maintains consistent attention to anomalies of varying degrees within the same sample.

**Visualizations on MAD-sys.** Compared to the weak defects selected from the test set of MVTec AD, the synthetic weak defects are even difficult to discover by the human eyes. Fig. S14 indicates that GLASS can still distinguish weaker synthetic defects even when other methods fail to localize any anomalies. Additionally, as the $\beta$ increases and defects become less pronounced, a noticeable performance decline in GLASS is only observed under $\beta = 0.7$. Fig. S15 demonstrates the detection results of the weak defects synthesized from 5 texture categories of MVTec AD under $\beta = 0.5$. Although GLASS achieves lower precision relative to MAD-man, it significantly surpasses all other methods across the five categories of MAD-sys.

Figure S14. Visualizations of GLASS-m compared with different SOTA methods on the 'Carpet' class of MAD-sys under different $\beta$. The ground truth is marked by the red boundary on the input.



\end{document}